\documentclass[10pt,twocolumn,letterpaper]{article}

\usepackage[pagenumbers]{cvpr} 

\usepackage{multirow} 
\usepackage{makecell} 

\newcommand{\argmin}[1]{\underset{#1}{\operatorname{arg}\,\operatorname{min}}\;}

\definecolor{cvprblue}{rgb}{0.21,0.49,0.74}
\usepackage[pagebackref,breaklinks,colorlinks,citecolor=cvprblue]{hyperref}

\def\paperID{*****} 
\def\confName{CVPR}
\def\confYear{2024}

\title{Semantic Guided Large Scale Factor Remote Sensing Image Super-resolution with Generative Diffusion Prior}

\author{Ce Wang, Wanjie Sun\footnote{Corresponding author}\\
School of Remote Sensing and Information Engineering, Wuhan University\\ Wuhan 430079, China\\
\tt\small {\{cewang, sunwanjie\}@whu.edu.cn}
}
\begin{document}
\maketitle
\begin{abstract}
Remote sensing images captured by different platforms exhibit significant disparities in spatial resolution. Large scale factor super-resolution (SR) algorithms are vital for maximizing the utilization of low-resolution (LR) satellite data captured from orbit. However, existing methods confront challenges in recovering SR images with clear textures and correct ground objects. We introduce a novel framework, the Semantic Guided Diffusion Model (SGDM), designed for large scale factor remote sensing image super-resolution. The framework exploits a pre-trained generative model as a prior to generate perceptually plausible SR images. We further enhance the reconstruction by incorporating vector maps, which carry structural and semantic cues. Moreover, pixel-level inconsistencies in paired remote sensing images, stemming from sensor-specific imaging characteristics, may hinder the convergence of the model and diversity in generated results. To address this problem, we propose to extract the sensor-specific imaging characteristics and model the distribution of them, allowing diverse SR images generation based on imaging characteristics provided by reference images or sampled from the imaging characteristic probability distributions. To validate and evaluate our approach, we create the Cross-Modal Super-Resolution Dataset (CMSRD). Qualitative and quantitative experiments on CMSRD showcase the superiority and broad applicability of our method. Experimental results on downstream vision tasks also demonstrate the utilitarian of the generated SR images. The dataset and code will be publicly available at https://github.com/wwangcece/SGDM
\end{abstract}
\section{Introduction}
High-resolution (HR) remote sensing imagery is crucial in various applications, such as weather prediction \citep{venter2020hyperlocal}, agricultural crop type identification \citep{sun2019using}, geographic object detection \citep{li2020object}, and land use/cover mapping \citep{yin2021integrating, tong2020land}. However, accessing HR images is often challenging due to limited satellite coverage, high costs, infrequent revisit times, reduced spectral resolution, and sensitivity to weather conditions \citep{liao2023high, bamford2020comparison}. In contrast, satellites with lower spatial resolution offer wider coverage, more frequent revisit, and better cost-effectiveness, but it lack critical details for precise object identification or monitoring. Consequently, there is a strong demand for software-based algorithms to enhance low-resolution (LR) images, providing an efficient and cost-effective means to leverage globally available LR remote sensing data.

Image super-resolution (SR) aims to reconstructs higher-resolution images with enhanced details from LR sources, enabling the acquisition of HR images beyond the sampling constraints of digital imaging system. As a result, algorithmic SR methods have become a prominent research topic in computer vision and image processing. Over the past decades, various SR approaches have been developed, including reconstruction-based \citep{sun2008image}, example-based \citep{freeman2002example}, sparse representation-based \citep{yang2010image}, and regression-based \citep{timofte2013anchored} approaches. With the advancement of deep learning techniques, deep neural network-based SR models have significantly outperformed traditional interpolation and optimization techniques. Early deep learning research in SR focused on the design of network architectures \citep{dong2015image, lai2017deep}, loss function \citep{johnson2016perceptual, bulat2018super}, and optimization strategy \citep{wang2018fully, lim2017enhanced}, leading to notable improvements in fidelity oriented quantitative performance. Recently, generative models like variational autoencoders \citep{liu:2021photo}, generative adversarial networks \citep{wang2018esrgan}, normalizing flows \citep{lugmayr2020srflow}, and diffusion models \citep{saharia2022image} have showcased impressive perceptual quality in SR results.

Previous research on SR in both natural and remote sensing images has predominantly focused on small scale factors, like 2$\times$ or 4$\times$. However, in real-world scenarios, there is often a significant disparity in spatial resolution between different remote sensing platforms. For instance, Landsat series satellites have a spatial resolution of approximately 80 meters in the visible and near-infrared bands, Sentinel series satellites have a spatial resolution of about 10 meters, and WorldView-3 satellites achieve a spatial resolution of 1.24 meters. Consequently, large scale factor SR models are essential to leverage LR satellite images for specific applications. However, large scale factor SR poses significant challenges due to the severe loss of detailed information in LR images, making SR a highly ill-posed problem with a vast solution space. To address this, researchers often employ prior knowledge from natural HR images to restrict SR results to a low-dimensional manifold. Commonly used priors include statistical models \citep{shaham2019singan}, sparse representation \citep{yang2010image}, and low-rank structures \citep{sun2022learning}. 

Recently, learning-based priors based on pre-trained neural networks, like PULSE \citep{menon2020pulse} and GLEAN \citep{chan2021glean}, have shown remarkable performance in large scale factor SR. The Stable Diffusion model, trained on billions of text-image pairs, has been employed as a generative prior in image restoration tasks, significantly enhancing visual quality by producing natural-looking, high-quality images \citep{wang2023exploiting, lin2023diffbir, yang2023pixel}. Nonetheless, employing a generative prior for large scale factor SR often leads to the hallucination of image details that do not align with the ground truth. As a result, harnessing pre-trained generative diffusion models as natural HR priors for generating remote sensing images with accurate structures and realistic textures remains a challenging task.

\begin{figure}[tp]
	\centering
        \includegraphics[width=\linewidth]{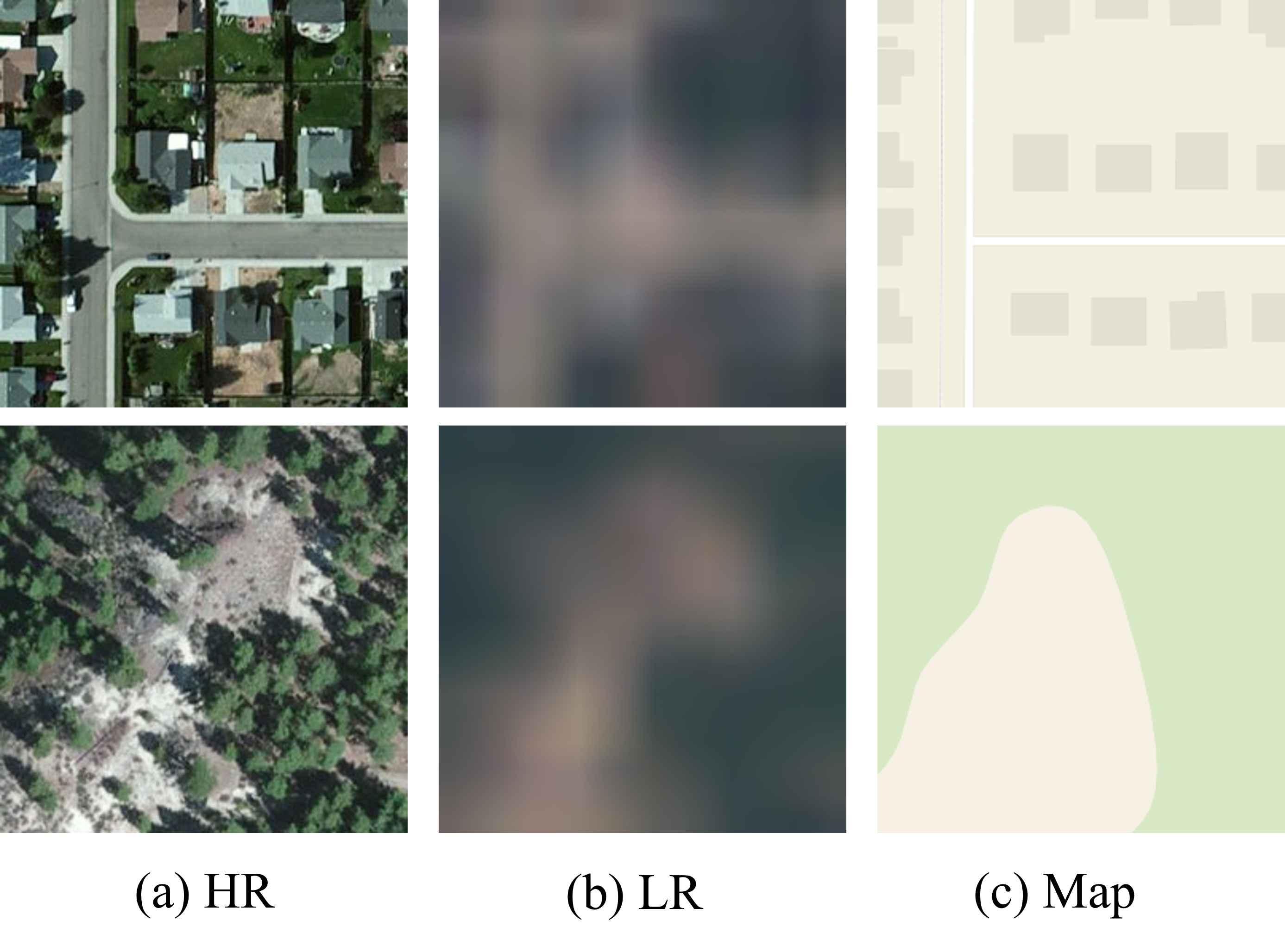}
	\caption{In real-world scenarios, there exists a significant spatial resolution disparity among remote-sensing images captured by different sensors. Vector maps, on the other hand, provide rich semantic guidance for large scale factor super-resolution.}
    \label{fig:data-sample}
\end{figure}

In Fig. \ref{fig:data-sample}, we showcase two pairs of LR and HR remote sensing images from a real-world scenario. The LR images are captured by Sentinel-2, with a resolution of 10 meters, while the HR images are sourced from the World Imagery, with a spatial resolution of 1.07 meters. The main challenges in large scale factor SR in real-world contexts stem from the significant difference between LR and HR images, including:
\begin{enumerate}[label=\textbullet]
    \item Difficulty in recovering semantically accurate and texture clear results. Due to the loss of most high-frequency information in LR images, it is challenging to reconstruct perceptual plausible and semantic correct HR images from LR images even with the state-of-the-art generative methods.
    \item Instability in model training due to pixel misalignment between LR and HR images. Due to the differences in imaging characteristics among different sensors, LR and HR images are aligned only at the semantic level rather than pixel level, posing significant challenges for stable model training.
\end{enumerate}

To circumvent the aforementioned issues, we propose a new framework named Semantic Guided Diffusion Model (SGDM) for large scale factor remote sensing image super-resolution. SGDM utilizes the pre-trained Stable Diffusion model to generate SR results with more realistic textures. However, relying solely on pre-trained generative models as a prior cannot produce structural and semantic accurate ground objects when SR factor is large. Fortunately, vector maps, which are readily available, provide strong semantic guidance for large scale factor SR tasks. Our proposed SGDM can effectively exploit vector maps for enhanced fidelity in the SR process. Furthermore, to accommodate the differences in imaging characteristics across various sensors in real-world scenarios, we propose a method to implicitly decouple imaging characteristics from image content, enabling probabilistic modeling of the imaging characteristics. Finally, we also introduce a new dataset, the Cross-Modal Super-Resolution Dataset (CMSRD), to validate and evaluate the proposed SGDM. Experimental results demonstrate that our method outperforms existing state-of-the-art (SOTA) methods in quantitative metrics and achieves greater flexibility and visual style control. Experimental results on downstream vision tasks, such as semantic segmentation and scene recognition, also demonstrate the effectiveness of the proposed SGDM and the usability of the generated SR results.

The main contributions of this work are summarized as follows: 
\begin{enumerate}[label=\textbullet]
    \item A novel remote sensing image super-resolution framework, named Semantic Guided Diffusion Model is proposed. It can achieve up to 32$\times$ super-resolution with accurate structures and realistic textures.
    \item We propose to incorporate vector maps into the super-resolution process, and validate its effectiveness for large scale factor super-resolution tasks by providing structural and semantic cues.
    \item We propose to decouple imaging characteristics from image content, addressing pixel-level inconsistencies between LR and HR images, and achieve diverse SR results generation with either imaging characteristics guidance or sampling.
    \item We provide an open-sourced Cross-Modal Super-Resolution Dataset (CMSRD), which contains 45300 pairs of synthetic LR/HR images and 50754 pairs real-world LR/HR images, to evaluate the proposed SGDM and facilitate the future research on large scale factor remote sensing image super-resolution for the community.
\end{enumerate}

The remainder of this paper is organized as follows. Section \ref{sec:Related work} reviews the work of remote sensing image super-resolution and large scale factor image super-resolution. In Section \ref{sec:Methodology}, we introduce the details of our method, including the main idea and the specific architecture of the semantic guided diffusion model. Section \ref{sec:Datasets and assessment metrics} introduces the data and evaluation metrics of this work. We present experimental results and ablation studies in Section \ref{sec:Experiments} and Section \ref{sec:Ablation studies}. And finally, Section \ref{sec:Conclusion} presents our conclusions.
\section{Related work}\label{sec:Related work}
\subsection{Remote sensing image super-resolution}\label{sec:Remote sensing image super-resolution}
In recent years, the rapid advancement of deep learning has led to the emergence of numerous deep learning-based super-resolution (SR) methods, showcasing their superiority over traditional techniques. These methods could be categorized into four main types: PSNR-oriented, GAN-based, flow-based, and diffusion-based methods. PSNR-oriented models \citep{dong2015image, lim2017enhanced, zhang2018image} are trained using straightforward loss functions (e.g., L1 or L2 loss), resulting in impressive PSNR scores. However, these loss functions tend to guide the SR results towards an average of several potential predictions, leading to over-smoothing images and the loss of high-frequency information. GAN-based approaches \citep{ledig2017photo, wang2018esrgan, zhang2021designing} address this issue by incorporating content (L1 or L2) and adversarial losses, enhancing perceptual quality but may struggle with convergence and mode collapse. Flow-based methods \citep{lugmayr2020srflow, liang2021hierarchical} use invertible encoders to tackle the ill-posed problem, but face high training costs due to strict architectural requirements. Recently, denoising diffusion probabilistic models (DDPM) \citep{ho2020denoising} have received increasing attention in the realm of image-to-image translation \citep{yang2023diffusion}, and also achieved promising performance in super-resolution tasks \citep{li2023diffusion}. SR3 \citep{saharia2022image} is the first to adapt DDPM for SISR tasks, yielding competitive perceptual quality. Although DDPM-based methods can model more intricate distributions and alleviate the training instability which are often encountered in GANs, these methods all involve training the diffusion models from scratch, which demands substantial computational resources and carries the potential risk of compromising the generative priors captured in generative models \citep{wang2023exploiting}.

Due to the rapid advancement of deep learning, deep learning-based remote sensing image SR research has also advanced. \cite{liebel2016single} was among the first to employ convolution neural networks for this task, outperforming traditional methods in upscaling Sentinel-2 images. Subsequently, studies on remote sensing image SR have primarily focused on two directions. The first direction involves designing specialized models for remote sensing images. Taking into account the complex edge details commonly found in remote sensing images, \cite{yang2017deep} devised an edge enhancement network and introduced edge loss. In light of much more complex structure of remote sensing images than that of natural images, \cite{zhang2020scene} proposed a scene-adaptive super-resolution strategy to more accurately describe the structural features of different scenes. In addition to improvements in super-resolution models, another direction aims to bridge the gap between synthetic and real-world imagery. For instance, \cite{zhu2020super} proposed a realistic training data generation model for commercial satellite imagery products and \cite{pineda2020generative} focused on enhancing the spatial resolution of Sentinel-2 satellite images using very high-resolution (VHR) PeruSat-1 images as a reference set.

Recently, the pre-trained Stable Diffusion model \citep{rombach2022high} has demonstrated success in content generation \citep{zhang2023adding}. As a result, there is growing interest in leveraging pre-trained models for super-resolution tasks \citep{wang2023exploiting, lin2023diffbir, yang2023pixel}. However, due to the distinct nature of remote sensing imagery, there has been limited exploration of integrating the pre-trained Stable Diffusion model into remote sensing image super-resolution. Despite this, the potential of the pre-trained model in this field remains substantial.

\subsection{Large scale factor image super-resolution}\label{sec:Large scale factor image super-resolution}
Due to the highly ill-posed nature of large scale factor super-resolution, many methods require additional prior information to achieve better outcomes. Methods based on explicit priors rely on additional inputs to provide more realistic texture or semantic guidance, such as reference images \citep{sun2022learning, zhang2020texture}, semantic masks \citep{buhler2020deepsee}, hyperspectral data \citep{meng2022large}, land cover change masks \citep{dong2024building}, etc. By aligning the input LR images and reference information in the feature space, these methods exploit the reference information to enhance super-resolution. However, acquiring high-quality references and precise feature alignment can be challenging in real-world situations, prompting researchers to explore implicit prior-based methods.

These methods typically utilize pre-trained GAN models to generate more realistic textures. \cite{menon2020pulse} introduced an unsupervised super-resolution approach based on pre-trained GANs, exploring the manifold of natural images to find HR images most similar to the LR input. \cite{gu2020image} employed multiple initial latent codes to generate intermediate feature maps and then compose them through channel attention mechanisms. In contrast to previous GAN inversion methods, \cite{chan2021glean} leveraged pre-trained StyleGAN \citep{karras2019style, karras2020analyzing} model and achieved visually pleasing reconstruction results with a single forward pass. However, such methods can only reconstruct images for specific categories (such as faces, cats, bedrooms, etc.) and struggle to generalize to generic scenes. 

In contrast to existing methods, our approach integrates both explicit priors (vector maps) and implicit priors (pre-trained Stable Diffusion model) to generate more realistic textures while maintaining semantic accuracy of the reconstructed results.
\section{Methodology}\label{sec:Methodology}
\subsection{Main idea}\label{sec:Main idea}
\begin{figure}[tp]
    \centering
    \includegraphics[width=\linewidth]{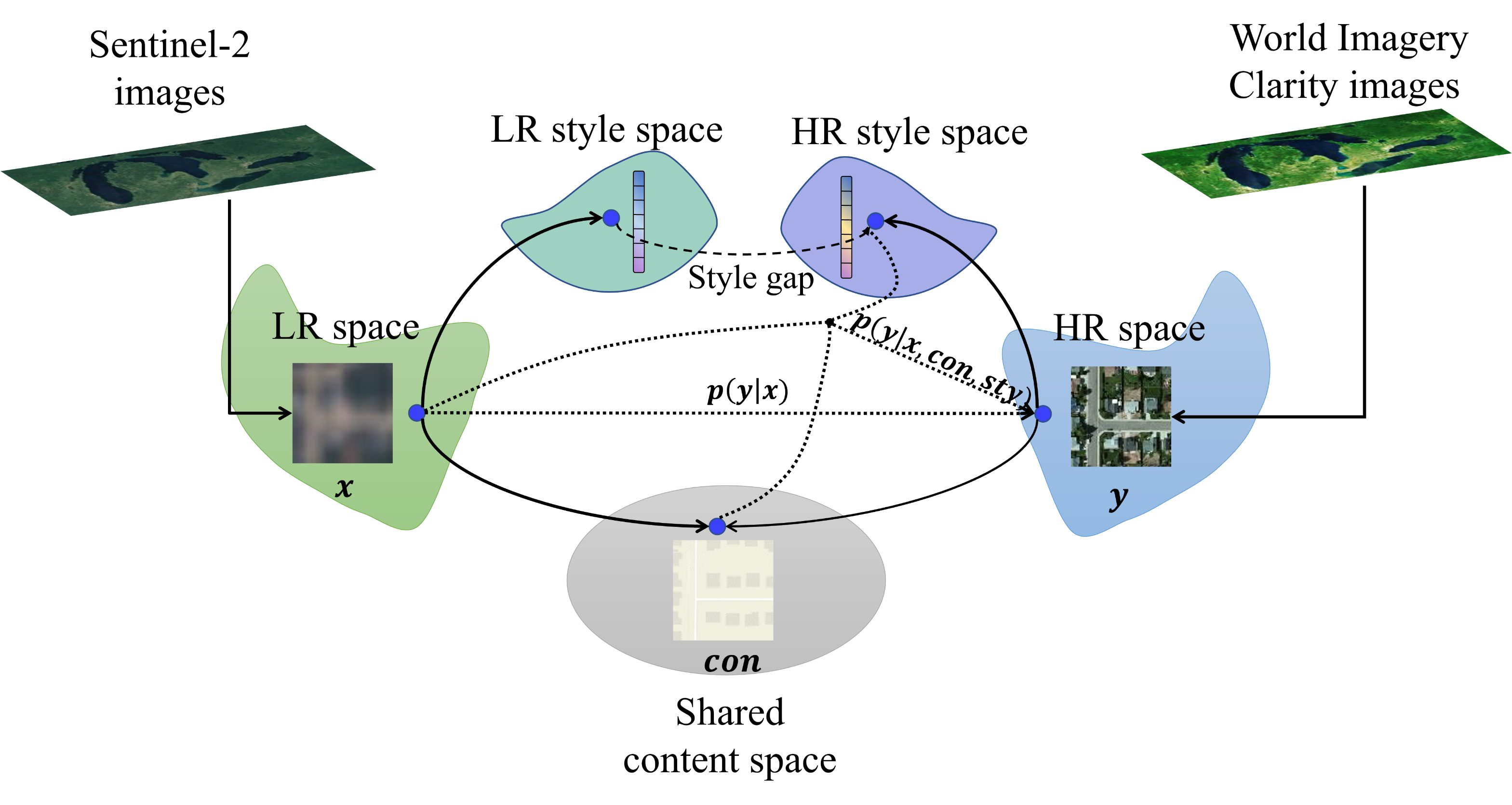}
    \caption{Main idea of our work. Unlike traditional super-resolution methods that only use LR, our approach additionally incorporates content conditions and style conditions. Content guidance can be provided by vector maps, while style guidance can be achieved through style-guided images or style sampling from HR style space.}
    \label{fig:motivation}
\end{figure}

The main idea of our work is illustrated in Fig. \ref{fig:motivation}. Traditionally, generative models for super-resolution only take LR images as input to produce SR images, which can be mathematically expressed as:
\begin{equation}
    \begin{aligned}
    y=\argmin{y}-\log\left(p\left(y\mid x\right)\right)
    \end{aligned}
    \label{equ:r1_ddpm_model}
\end{equation}
where $x$ represents the LR image, $y$ represents the estimated SR image. $p(\cdot\mid x)$ is the conditional probability density function of HR images given LR inputs, which is usually modelled as the standard Gaussian (L2 loss) or Laplacian (L1 loss) distribution with the ground truth HR images as mean. Theoretically, this modeling approach that relies solely on LR images can learn the many-to-one mapping between LR and HR images. However, these methods often struggle to obtain sufficiently accurate semantic information about ground objects when the SR factor is large. Additionally, the differences in imaging characteristics among different sensors can hinder the model from converging to its optimal state, resulting in the model to produce the mean value of all possible SR results.

Considering the issues mentioned above and inspired by research on disentangled representation learning, our work proposes the following modeling approach:
\begin{equation}
    \begin{aligned}
    y=\argmin{y}-\log\left(p\left(y\mid x,con,sty\right)\right)
    \end{aligned}
    \label{equ:r1_our_model}
\end{equation}
where $con$ and $sty$ represent the content condition and style condition respectively. In our work, the terms `content' and `style' have the following definitions: the content component of an image refers to its semantics, which represents the structure of ground objects within the image, whereas the style component refers to a representation of the sensor-specific imaging characteristics. Any useful prior can be used to provide content guidance during training, and diverse reconstruction results can be obtained through style guidance or style sampling during testing. By decomposing the image into content and style components, this modeling approach enables better guidance of the reconstruction results with higher fidelity and plausible diversity.

\subsection{Framework overview}\label{sec:Main idea and framework overview}
\subsubsection{Preliminary: diffusion model}
Denoising Diffusion Probabilistic Models (DDPM) are a class of generative models based on the diffusion process. The core idea of DDPM is to simulate the diffusion process of data by gradually adding noise and then progressively removing the noise to create new data samples \citep{ye2023ip}. These two processes are referred to as forward diffusion and reverse denoising, which correspond to the training and sampling procedures of the model. The classic DDPM executes forward diffusion and reverse denoising processes across the entire pixel space with the diffusion process mathematically described as:
\begin{equation}
    \begin{aligned}
    y_{t} = \sqrt{\bar{\alpha_{t}} } y_{0}+\sqrt{1-\bar{\alpha_{t}}}\epsilon, \ \ \ \ \epsilon\sim N\left(0,1\right)
    \end{aligned}
    \label{equ:r1_diffusion_process}
\end{equation}
where $y_{0}$ denotes images sampled from the true data distribution, $\bar{\alpha_{t}}$ represents a pre-set hyperparameter related to noise patterns. The term $\epsilon$ refers to random noise drawn from a standard Gaussian distribution. DDPM then focuses on training a denoiser function, $\epsilon_{\theta}$, which aims to estimate the noise added to $y_{t}$, subject to an $L_{2}$ loss constraint:
\begin{equation}
    \begin{aligned}
    L=\mathrm{\mathbb{E}}_{\left(y_{0},\epsilon, t\right)}\parallel \epsilon-\epsilon_{\theta}\left(y_{t},t\right) \parallel ^{2}
    \end{aligned}
    \label{equ:r1_l2_loss}
\end{equation}
where $t$ represents an integer timestamp sampled from the interval $[0, T]$ and $\epsilon_{\theta}$ is the noise predictor that we need to train. Following the optimization of the model based on Equation \ref{equ:r1_l2_loss}, high-quality images can be produced by sequentially refining random noise through the denoising process. If we want to control the generation process more precisely, additional conditional variables can be added to the noise predictor, \ie, $\epsilon_{\theta}\left(y_{t},t, c\right)$ in Equation \ref{equ:r1_l2_loss}. In accordance with the principles of DDPM, each step of this denoising process can be mathematically described as:
\begin{equation}
    \begin{aligned}
    y_{t-1}=\frac{1}{\sqrt{\alpha_{t}}}(y_{t}-\frac{\beta_{t}}{\sqrt{1-\bar{\alpha_{t}}}}\epsilon_{\theta})+\sigma_{t}\epsilon, \ \ \ \ \epsilon\sim N\left(0,1\right)
    \end{aligned}
    \label{equ:r1_denoising_process}
\end{equation}
where $\alpha_{t}$, $\bar{\alpha _{t} }$, $\beta_{t}$, and $\sigma_{t}$ are hyperparameters related to noise patterns, and $\epsilon$ represents additive noise added during the inverse process. After iterating $T-1$ times, we can obtain the results sampled under the condition $c$.

However, classical DDPMs often demand substantial computational resources and time for both training and inference, limiting their broader application. In contrast, latent diffusion models shift the diffusion and denoising steps from pixel space to the latent space of a pre-trained autoencoder with much smaller spatial resolution. The encoding and decoding processes of the autoencoder can be expressed as follows:
\begin{align}
    z_{0} = \mathcal{E}\left( y \right)
    \label{equ:r1_encoder} \\
    y \approx \mathcal{D}\left( z_{0} \right)
    \label{equ:r1_decoder}
\end{align}
where $z_{0}$ denotes the latent representation of the HR image $y$, $\mathcal{E}$ denotes the pre-trained encoder, and $\mathcal{D}$ represents the pre-trained decoder. Then the entire diffusion and denoising process will subsequently take place in the latent space where $z_{0}$ resides, thereby achieving an optimal balance between reducing complexity and preserving details. 

\begin{figure*}[htp]
    \centering
    \includegraphics[width=\linewidth]{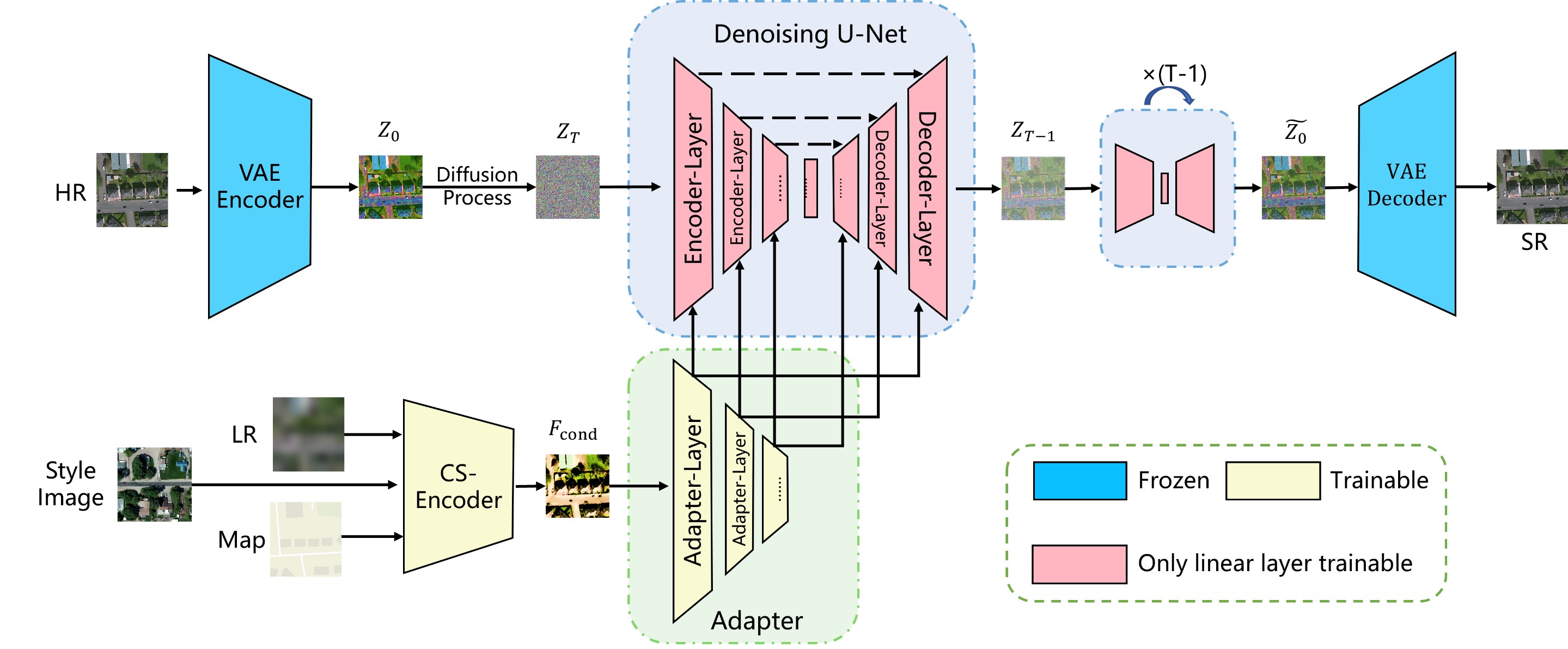}
    \caption{The framework of SGDM. A VAE is used to shift the diffusion process and reverse process from the pixel space to the latent space. During training, the latent features $z_{0}$ of HR are transformed into $z_{t}$ through the diffusion process, which is then denoised through a U-Net network. To guide the denoising process, we design two modules: the content-style encoder (CS-Encoder) and the adapter. The former integrates information from LR, vector maps, and style guidance images to generate conditional features $F_{\mathrm{cond}}$. The latter generates multi-scale features based on $F_{\mathrm{cond}}$ and performs element-wise addition with the output of the corresponding layers in the U-Net.}
    \label{fig:r3_framework}
\end{figure*}

\subsubsection{Semantic guided diffusion model}
Based on the discussion of challenges in large scale factor super-resolution of remote sensing images in real-world scenarios, we propose a new super-resolution framework called Semantic Guided Diffusion Model (SGDM). The main structure of the SGDM is shown in Fig. \ref{fig:r3_framework}. Unlike previous single-image super-resolution frameworks, SGDM can simultaneously accept inputs of LR images, high-resolution style guidance images, and vector maps to jointly reconstruct HR images.

The entire framework of the SGDM can be divided into four parts: a pre-trained variational autoencoder (VAE), a denoising U-shaped network (U-Net) with skip connections, a content-style encoder (CS-Encoder), and an adapter. The VAE is used to encode and decode HR images, thereby moving the diffusion and denoising processes of classical DDPM from the pixel space to the latent space, which enhances the stability of the training process and reduces computational resource consumption. The U-Net is a conditional noise predictor that iteratively predicts noise to obtain noise-free data. It consists of several encoding layers, middle layers, and decoding layers, most of which are composed of residual modules, cross-attention modules, and self-attention modules. The content-style encoder merges information from the style guidance image, LR image, and vector map, thus generating a conditional feature map encapsulating semantic and style information. Lastly, the adapter translates this conditional feature into multi-scale features compatible with the U-Net network architecture. These features are then element-wise added to the encoding and decoding layers of the U-Net, ensuring effective feature fusion and guidance during the generation process.

\subsection{Content-style encoder}\label{sec:Content-style encoder}
\begin{figure*}[htp]
    \centering
    \includegraphics[width=\linewidth]{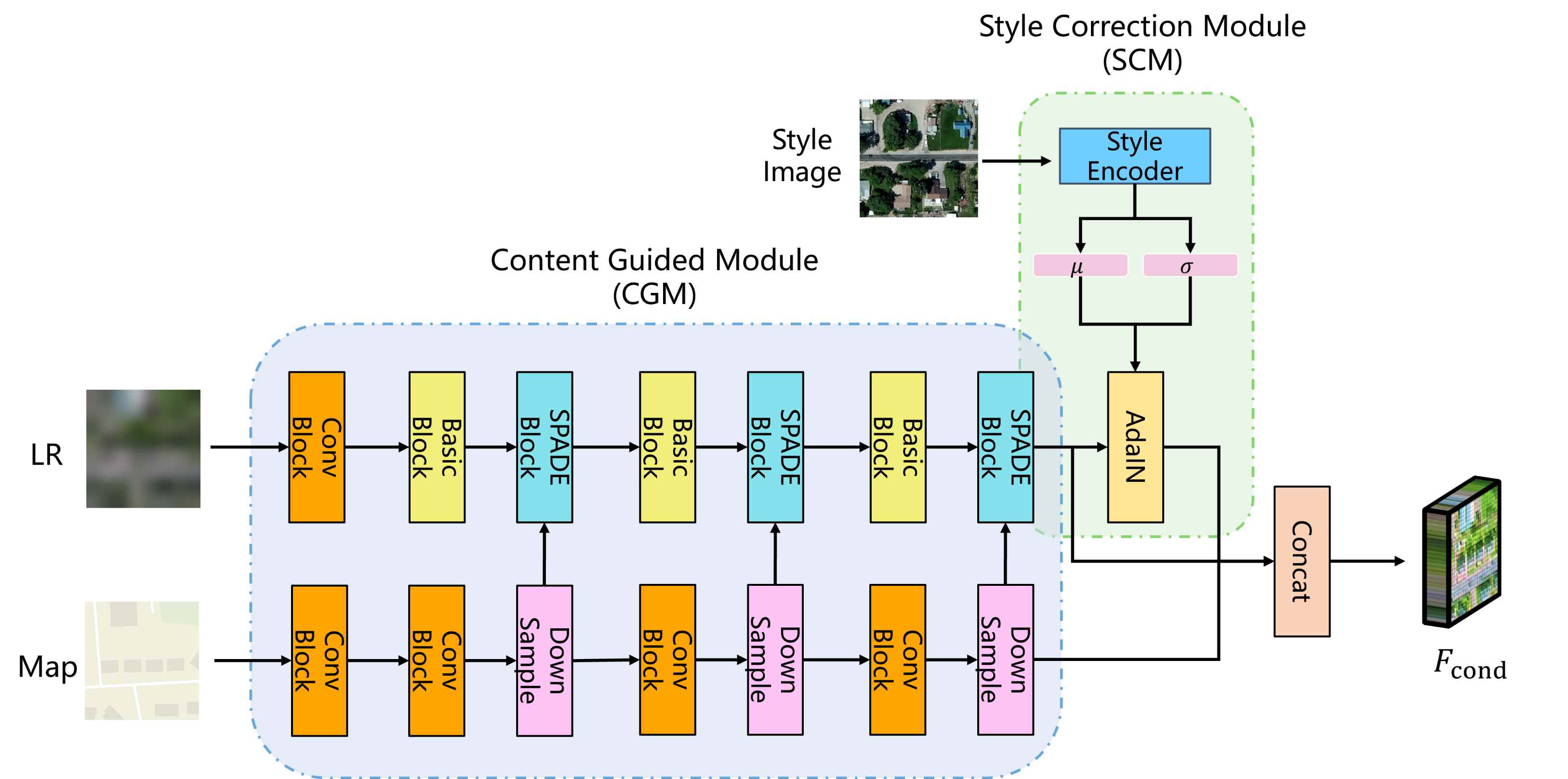}
    \caption{Detailed structure of the proposed Content-Style Encoder (CS-Encoder). It can utilize the content information from the vector map at multiple scales through the SPADE module, and achieve style injection through the AdaIN module, which results in a conditional feature $F_{\mathrm{cond}}$. This feature encapsulates semantic information and style attributes.}
    \label{fig:r3_sc_encoder}
\end{figure*}

To efficiently fuse content information from vector maps and inject style information from style guidance images into LR images, a multi-branch content-style encoder (CS-Encoder) is designed, whose detailed structure is shown in Fig. \ref{fig:r3_sc_encoder}. It consists two key components: the content guided module (CGM) and the style correction module (SCM), which together produce a conditional feature with semantic information and style information.

\subsubsection{Content guided module}\label{sec:Content guided module}
\begin{figure}[tp]
    \centering
    \includegraphics[width=\linewidth]{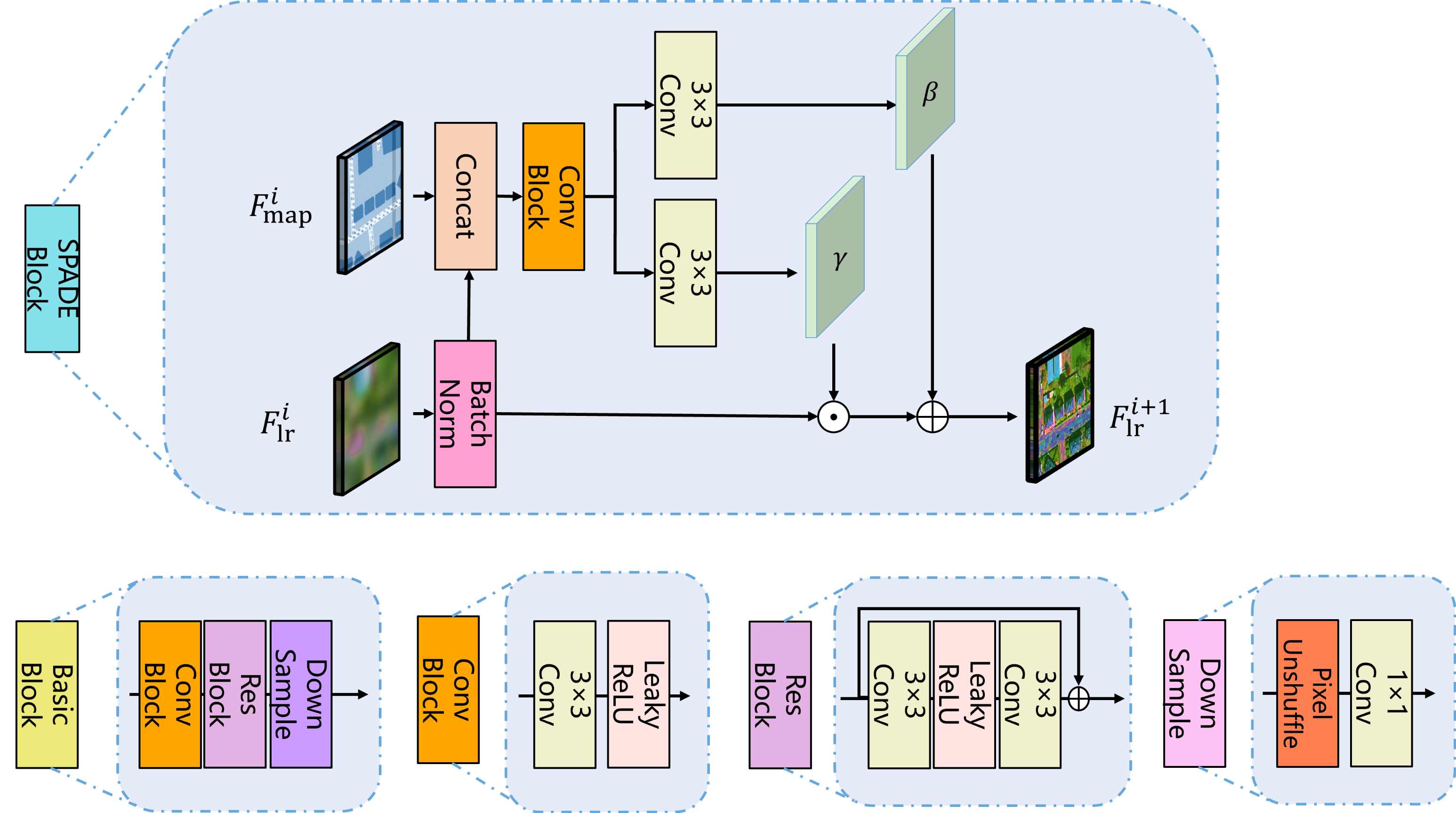}
    \caption{Structures of SPADE block, basic block and downsample block in our work. }
    \label{fig:modules}
\end{figure}

In the content guidance module, we integrate semantic information from vector maps into LR images at multiple scales. In the first branch, multi-scale features of the vector map are extracted through convolution and downsample blocks. In another branch, multi-scale features of the LR image are extracted through the basic blocks. To fuse these features, we employ spatially-adaptive denormalization (SPADE) \citep{park2019semantic} modules at each scale. The network structures of the basic block, convolution block, downsample block, and SPADE block are illustrated in Fig. \ref{fig:modules}.  It is worth noting that in our design, the input of SPADE block is the feature map of LR images, which possesses very coarse texture and structural information. Consequently, our adaptation of the SPADE module differs from the original one by using both LR features and vector map features to calculate the scale factor $\gamma$ and bias term $\beta$. This enables more flexibility and higher-fidelity fusion of content guidance informatio into LR image features. The entire process is described by the following equation:
\begin{gather}
    \gamma^{i} = \mathrm{Conv}_{\gamma}\left(\mathrm{Conv_{share}}\left(\left[F_{\mathrm{lr}}^{i},F_{\mathrm{map}}^{i}\right]\right)\right)
    \label{equ:r1_gamma} \\
    \beta^{i} = \mathrm{Conv}_{\beta }\left(\mathrm{Conv_{share}}\left(\left[F_{\mathrm{lr}}^{i},F_{\mathrm{map}}^{i}\right]\right)\right)
    \label{equ:r1_beta} \\
    F_{\mathrm{lr}}^{i+1} = \gamma^{i} \frac{F_{\mathrm{lr}}^{i} - \mu^{i}}{\sigma^{i}} + \beta^{i}
    \label{equ:r1_spade}
\end{gather}
where $F_{\mathrm{lr}}^{i}$ and $F_{\mathrm{map}}^{i}$ represent the $i$-th layer of the LR image and the vector map features. $\mathrm{Conv_{share}}$ represents the shared convolution module, while $\mathrm{Conv}_{\gamma}$ and $\mathrm{Conv}_{\beta}$ represent convolution layers used to respectively predict scale factor $\gamma^{i}$ and bias $\beta^{i}$. $\gamma^{i}$ and $\beta^{i}$ represent the spatially variant scale factor and bias for the $i$-th layer, while $\mu^{i}$ and $\sigma^{i}$ represent the mean and standard deviation of all input features for the $i$-th layer within a mini-batch. By employing stacked content guided modules, the content information from the vector map is progressively integrated into LR features, resulting in more semantically accurate reconstructions.

\begin{figure}[tp]
    \centering
    \includegraphics[width=\linewidth]{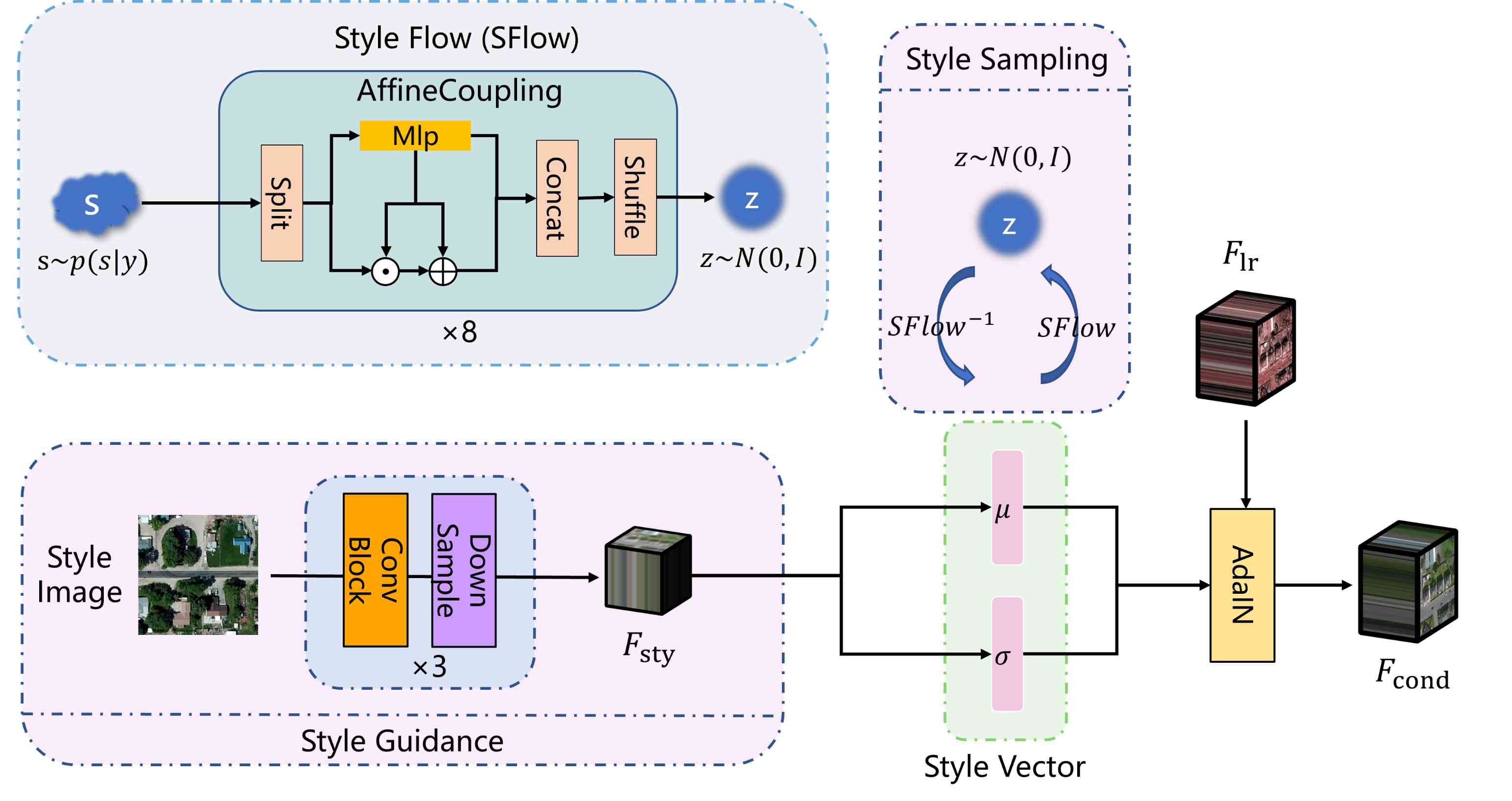}
    \caption{Structure of style correction module (SCM) proposed by us. It can achieve style guidance or style sampling through the style encoder or normalization flow respectively.}
    \label{fig:SIM}
\end{figure}

\subsubsection{Style correction module}\label{sec:Style correction module}
To overcome the style gap between LR and HR images during model training, we propose a simple yet effective style correction module (SCM). Drawing inspiration from style transfer tasks, our approach entails extracting style features with a lightweight style encoder, followed by the application of adaptive instance normalization (AdaIN) \citep{huang2017arbitrary}, as depicted in Fig. \ref{fig:SIM}. The AdaIN process can be mathematically represented as:
\begin{equation}
    \text{AdaIN}\left(F_{\mathrm{lr}}, F_{\mathrm{sty}}\right) = \sigma\left(F_{\mathrm{sty}}\right) \left( \frac{F_{\mathrm{lr}} - \mu(F_{\mathrm{lr}})}{\sigma(F_{\mathrm{lr}})} \right) + \mu\left(F_{\mathrm{sty}}\right)
    \label{equ:r1_adain}
\end{equation}
Here, $F_{\mathrm{lr}}$ represents the output of the content guided module, while $F_{\mathrm{sty}}$ represents the feature maps obtained by passing the style guidance image through the style encoder. $\mu(F_{\mathrm{lr}})$, $\sigma(F_{\mathrm{lr}})$, $\mu(F_{\mathrm{sty}})$, and $\sigma(F_{\mathrm{sty}})$ represent the mean and standard deviation computed across spatial dimensions independently for each channel and each sample. Numerous style transfer studies emphasize that channel-wise statistical features in image feature maps carry crucial style information \citep{ulyanov2016instance, li2017universal}. During training, we use HR images as style guidance, enabling the AdaIN module to incorporate HR image style into the reconstruction, thus enhancing content consistency learning and addressing the style discrepancy between LR and HR in real-world scenarios. During testing, any HR remote sensing images can be used as style guidance.

However, in real-world scenarios, obtaining effective style guidance images may be tedious. In such case, we resort to style sampling to obtain plausible stylized super-resolution results. Our proposed style correction module (SCM) not only supports style guidance but also facilitates style sampling. After training the entire model, we utilize the trained style encoder network to recompute the style vectors $\mu(F_{\mathrm{hr}})$ and $\sigma(F_{\mathrm{hr}})$ for all HR images in the training set. We then employ a normalizing flow model to model the probability distributions of these vectors separately (which corresponds directly to the term `HR style space' depicted in Fig. \ref{fig:motivation}). The style normalizing flow (SFlow) model, as illustrated in Fig. \ref{fig:SIM}, utilizes stacked affine coupling modules \citep{dinh2016density} to map input samples conforming to arbitrary complex probability distributions into standard Gaussian distributions. SFlow can be trained efficiently with exact log-likelihood estimation and enables random style sampling during testing, thus allowing for obtaining diverse plausible super-resolution results even without explicit style guidance images.

\subsection{Adapter design}\label{sec:Adapter design}
CS-Encoder can sufficiently integrate coarse texture and structural features from LR images with content features from vector maps and style information from style guidance images, ultimately producing a conditional feature $F_{\mathrm{cond}}$. Since $F_{\mathrm{cond}}$ already contains most of the external information required for super-resolution, the role of the adapter is merely to generate multi-scale features and align them with the prior knowledge of the pre-trained Stable Diffusion model. 

Therefore, the structure of the adapter module in Fig. \ref{fig:r3_framework} is relatively simple and it mainly consists of stacked convolution layers, residual blocks, and downsample blocks. The adapter module progressively increases the channel number of the feature maps while reducing their spatial sizes, thereby generating three different-sized conditional features. These features are then added element-wise to the output of each layer of the U-Net network for guiding the generation process of the Stable Diffusion model.

\subsection{Loss function}\label{sec:Loss function}
The training process of the Semantic Guided Diffusion Model (SGDM) comprises two stages. The first stage corresponds to training the SGDM, whereas the second stage corresponds to training the SFlow model. The trainable parameters of the first stage are indicated in Fig. \ref{fig:r3_framework}, and the loss function is as follows: 
\begin{equation}
    \begin{aligned}
    L=\mathrm{\mathbb{E}}_{\left(z_{0},\epsilon,lr,map,t\right)}\parallel \epsilon-\epsilon_{\theta}\left(z_{t},t,lr,map\right) \parallel ^{2}
    \end{aligned}
    \label{equ:r3_l2_loss}
\end{equation}
where $lr$ and $map$ denotes the input LR images and vector maps, $\epsilon$ refers to randomly sampled Gaussian noise, $t$ represents an integer timestamp sampled from the interval $[0, T]$ and $\epsilon_{\theta}$ is the denoising U-Net model.

After completing the first stage of training, we can recompute the style vectors $\mu(F_{\mathrm{lr}})$ and $\sigma(F_{\mathrm{lr}})$ for HR images in the training dataset and model their probability distributions using SFlow. The normalization flow model aims to establish a bijective mapping from an arbitrary distribution to a easy-to-sample distribution (such as the standard Gaussian distribution) through a series of chained invertible networks. Therefore, negative log-likelihood (NLL) can be used as the loss function to optimize it. Taking $\mu$ as an example, assuming $h_{k}$ represents the output of the $k$-th affine coupling layer $f_{\theta }^{k}$, and $h_{0}$ and $h_{N}$ represent the original input $\mu$ and the output random variable $z$, then the loss function of SFlow can be expressed as:
\begin{equation}
    \begin{split}
        \text{NLL} & = -\mathrm{log}~p\left(\mu\right) = \frac{D}{2}\mathrm{log}\left(2\pi\right)+\frac{1}{2}\left \| f_{\theta }\left(\mu\right) \right \|^{2} \\ 
        & - \sum_{k=1}^{N} \mathrm{log}\left | \mathrm{Det}\frac{\partial f_{\theta }^{k}\left(h^{k-1} \right) }{\partial h^{k-1}}   \right | 
        \label{equ:r3_flow_loss}
    \end{split}
\end{equation}
where $f_{\theta }$ represents the forward process of SFlow.

\begin{figure*}[htp]
    \centering
    \includegraphics[width=\linewidth]{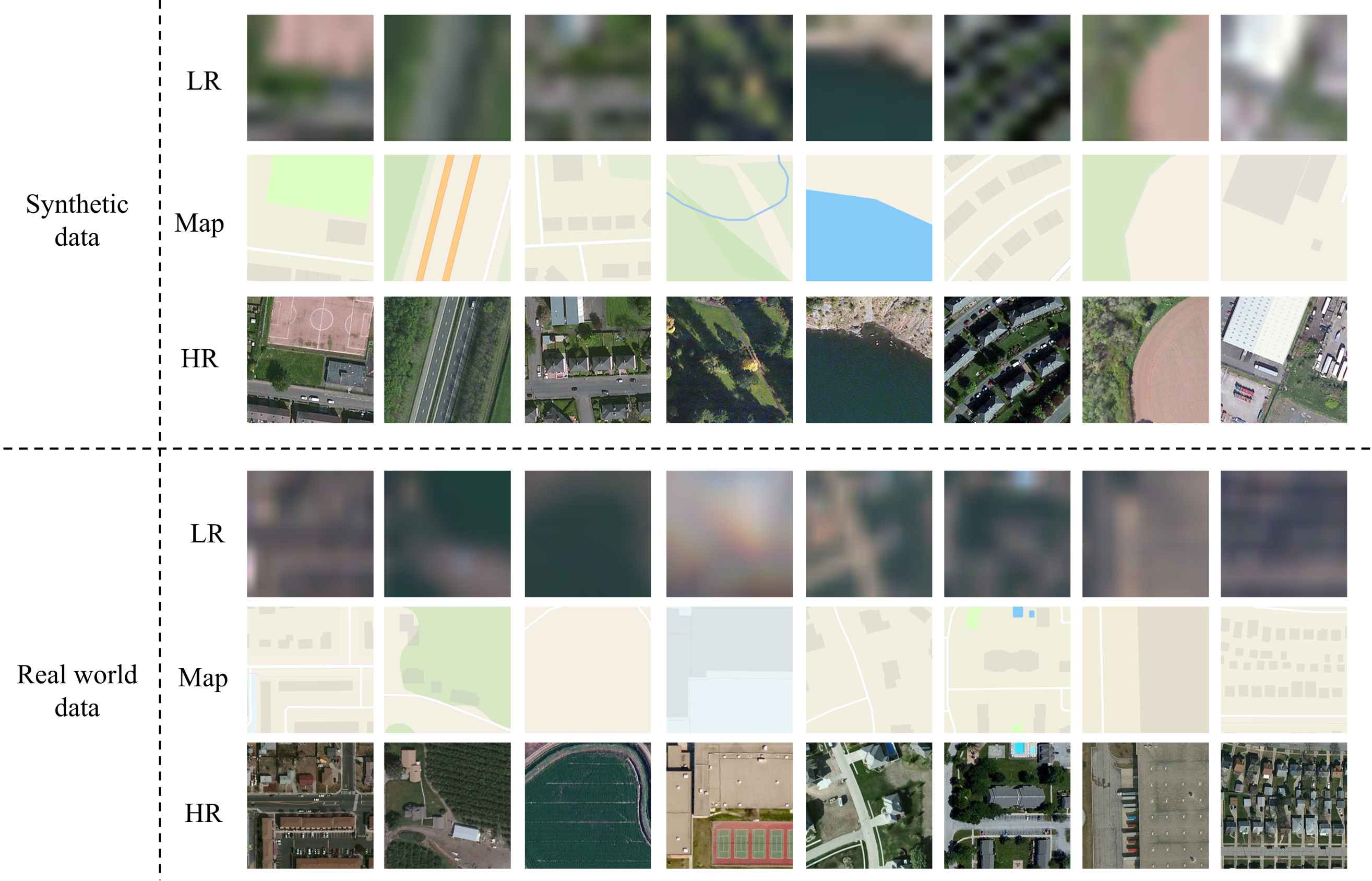}
    \caption{Examples of LR-Map-HR pairs in the CMSRD dataset. The dataset covers a wide range of scenes, including forests, rivers, streets, residential areas, industrial areas, water bodies, squares, and so on.
    }
    \label{fig:dataset_examples}
\end{figure*}

\begin{table*}
	\caption{Detailed information about the CMSRD.}
	\label{tab:cmsrd_dataset}
	\centering
	\begin{tabular}{cccccccc}
		\hline
         Type &Dataset &\makecell[c]{Image pair\\ number} &HR source &LR source &\makecell[c]{Vector map\\ source} &\makecell[c]{Resolution of\\ HR} &\makecell[c]{Resolution of\\ LR}\cr
        \hline 
        \multirow{2}{*}{Synthetic} 
        &Training set &45000 &World Imagery &Bicubic down &OSM &1.07 m &34.24 m\cr
        &Testing set &300 &World Imagery &Bicubic down &OSM &1.07 m &34.24 m\cr\hline
        
        \multirow{2}{*}{Real world} 
        &Training set &50364 &World Imagery  &Sentinel-2 &OSM &1.07 m &10 m\cr
        &Testing set &390 &World Imagery  &Sentinel-2 &OSM &1.07 m &10 m\cr\hline
	\end{tabular}
\end{table*}
\section{Datasets and image quality assessment metrics}\label{sec:Datasets and assessment metrics}
\subsection{Datasets}\label{sec:Datasets}
In this paper, we collect a new benchmark dataset named Cross-Modal Super-Resolution Dataset (CMSRD) to validate the proposed SGDM and hope to facilitate future research on large scale factor remote sensing image super-resolution. This dataset is composed of two parts: synthetic data and real-world data, with detailed information depicted in Table \ref{tab:cmsrd_dataset} and some examples of the image pairs shown in Fig. \ref{fig:dataset_examples}.

The CMSRD consists of triplets LR-Map-HR, all HR images are sourced from World Imagery, while all vector maps are obtained from Open Street Map. The synthetic dataset comprises 45,300 image pairs, with 45,000 pairs used for training and 300 pairs for testing. LR images are obtained by downsampling HR images by a factor of 32 using bicubic interpolation. The synthetic dataset is used to validate the effectiveness of introducing vector maps in large scale factor remote sensing image super-resolution. The real-world dataset consists of 50,754 image pairs, with 50,364 used for training and 390 for testing. In contrast to the synthetic dataset, LR images in the real-world dataset are sourced from the Sentinel-2 satellite series. We ensure 16$\times$ spatial size difference between HR and LR through bicubic resampling. As shown in Fig. \ref{fig:dataset_examples}, there exists content consistency and style gap among remote sensing images from different sources, which is one of the main issues to be explored in this work.

\subsection{Image quality assessment metrics}\label{sec:Assessment metrics}
To conduct a comprehensive evaluation of various remote sensing image super-resolution methods, we employ both full-reference and no-reference image quality assessment metrics. For assessing reconstruction fidelity, peak signal noise ratio (PSNR) and structure similarity (SSIM) \citep{wang2004image} calculated on the Y channel of the YCbCr color space are used. For the perceptual quality evaluation, full-reference metrics learned 
perceptual image patch similarity (LPIPS) \citep{zhang2018unreasonable} and deep image structure and texture similarity (DISTS) \citep{ding2020image} are employed. Furthermore, fréchet inception distance (FID) \citep{heusel2017gans} is used to gauge the distributional disparity between the ground truth images and the SR images. For no-reference image quality assessment, we use metrics such as multi-scale image quality transformer (MUSIQ) \citep{ke2021musiq} and CLIPIQA \citep{wang2023exploring}.

\begin{table*}
	\caption{Quantitative comparisons with different methods on the synthetic data for 32$\times$ SR and real-world data for 16$\times$ SR, where the bold font indicates the best performance. The symbols $\uparrow$ and $\downarrow$ respectively represent that higher or lower values indicate better performance.}
	\label{tab:quantitative comparison}
	\centering
	\begin{tabular}{ccccccccc}
		\hline
         Type &Metric &ESRGAN &SwinIR &LDM &ResShift &TTST &EDiffSR &SGDM/SGDM+\cr
        \hline
        \multirow{7}{*}{Synthetic} 
        &PSNR$\ \uparrow$ &20.26 &21.74 &20.12 &19.29 &\textbf{21.88} &20.32 &20.65\cr
        &SSIM$\ \uparrow$ &0.3233 &0.4029 &0.3121 &0.3021 &\textbf{0.4064} &0.2836 &0.3338\cr
        &LPIPS$\ \downarrow$ &0.4551 &0.8932 &0.4925 &0.4585 &0.8396 &0.4517 &\textbf{0.4344}\cr
        &DISTS$\ \downarrow$ &0.2561 &0.5528 &0.2609 &0.2537 &0.5001 &0.2480 &\textbf{0.2356}\cr
        &FID$\ \downarrow$ &70.67 &357.8 &42.43 &86.24 &295.3 &108.3 &\textbf{36.62}\cr
        &MUSIQ$\ \uparrow$ &38.13 &19.99 &39.40 &37.31 &17.77 &38.54 &\textbf{40.30}\cr
        &CLIPIQA$\ \uparrow$ &0.5951 &0.2683 &0.6116 &0.4546 &0.2471 &0.4590 &\textbf{0.6238}\cr \hline
        
        \multirow{5}{*}{Real world}
        &LPIPS$\ \downarrow$ &\textbf{0.5592} &0.9137 &0.6006 &0.6218 &0.9108 &0.6064 &0.5978\cr
        &DISTS$\ \downarrow$ &\textbf{0.3124} &0.5948 &0.3225 &0.3516 &0.5615 &0.3458 &0.3231\cr
        &FID$\ \downarrow$ &98.15 &423.2 &49.40 &166.4 &359.7 &156.1 &\textbf{44.81}\cr
        &MUSIQ$\ \uparrow$ &44.23 &20.51 &46.70 &44.41 &19.11 &42.21 &\textbf{48.92}\cr
        &CLIPIQA$\ \uparrow$ &0.4727 &0.2581 &0.5956 &0.3989 &0.2547 &0.4385 &\textbf{0.6064}\cr \hline
	\end{tabular}
\end{table*}
\section{Experiments}\label{sec:Experiments}
\subsection{Implementation details}\label{sec:Implementation details}
Our SGDM is built based upon the Stable Diffusion 2.1-base model. We implement the proposed SGDM using the PyTorch framework and all experiments are conducted on two NVIDIA GeForce RTX 3090 GPUs. 

The training phase consists of two stages. Firstly, we fine-tune the Stable Diffusion model using the loss function in Equation \ref{equ:r3_l2_loss}. During this phase, we optimize the parameters of content-style encoder, adapter, and the linear layers of U-net, which amounts to a total of 148M trainable parameters. We train the model with a batch size of 32 for 150K steps, and employ the AdamW optimizer with a learning rate of $5 \times 10^{-5}$. Subsequently, we utilize the trained style encoder to re-calculate style vectors ($\mu$ and $\sigma$) for all HR images in the training set and use them to train the SFlow model. The SFlow architecture follows the structure proposed in \cite{dinh2016density}, with hidden layer dimensions set to 16, 32, 32, 64 and the number of affine coupling layers set to 8. We employ two separate SFlow models for the distributions of $\mu$ and $\sigma$. We use the Adam optimizer with a learning rate of $4 \times 10^{-4}$, and the training of SFlow can be completed within two hours.

It is important to note that our style correction module (SCM), designed as a plug-and-play module, is not required when experimenting on synthetic datasets (referred to as SGDM). However, for real-world datasets, we integrate the SCM to address the style discrepancies between LR and HR images (referred to as SGDM+). Considering that previous methods for large scale factor super-resolution are only suitable for specific types of scenes, such as face \citep{chan2021glean, gu2020image, menon2020pulse}, here, we select some state-of-the-art (SOTA) SR methods for natural images to do the comparison, such as ESRGAN \citep{wang2018esrgan}, SwinIR \citep{liang2021swinir} and ResShift\citep{yue2024resshift}. LDM \citep{rombach2022high} is also chosen as a comparison method to validate the effectiveness of the semantic and style guidance we proposed. In addition, we also select two SOTA methods for remote sensing image super-resolution for comparison, \ie, the EDiffSR \citep{xiao2023ediffsr} and TTST \citep{xiao2024ttst}. For the sake of fairness, all methods are trained on our CMSRD dataset using the source code and hyper-parameter settings provided by those papers. It should be noted that the original hyper-parameter settings of ESRGAN results in convergence issues in scenarios of large scale factor super-resolution. Therefore, we adjust the scale of its loss functions to make it better suited for this task.

\subsection{Experiments on synthetic data}\label{sec:Experiments on synthetic data}
\subsubsection{Quantitative SR results}\label{sec:Quantitative results(syc)}

Table \ref{tab:quantitative comparison} presents 32$\times$ SR results of different methods compared on synthetic data. Transformer-based regression models, such as SwinIR and TTST, exhibit the highest PSNR and SSIM scores. However, these methods often produce over smooth SR images, which negatively impacts perceptual quality, as reflected by the low LPIPS, DISTS, FID, MUSIQ, and CLIPIQA scores. Meanwhile, ESRGAN, ResShift, and EDiffSR, as generative models based on GAN, diffusion, and SDE, exhibit similar performance. They show improved perceptual quality but still struggle for maintaining fidelity, leading to lower PSNR and SSIM. However, our proposed SGDM outperforms in both aspects and achieves the best balance between perceptual quality and fidelity. In addition to achieving the best performance across all perceptual metrics, SGDM also achieves relatively high PSNR and SSIM scores. On one hand, SGDM can utilize coarse texture and structural information from LR images and explicit semantic information from vector maps to effectively improve the fidelity of the reconstruction results; on the other hand, by leveraging the powerful generative prior of the Stable Diffusion, SGDM is still able to recover detailed textures that match real scenes even in extreme large factor super-resolution scenarios. Notably, despite relatively good perceptual metrics of LDM, it has a significantly higher number of trainable parameters (865M) compared to SGDM (148M), which highlights the efficiency and effectiveness of our network architecture.

\subsubsection{Qualitative SR results}\label{sec:Qualitative results(syc)}
\begin{figure*}[htp]
    \centering
    \includegraphics[width=\linewidth]{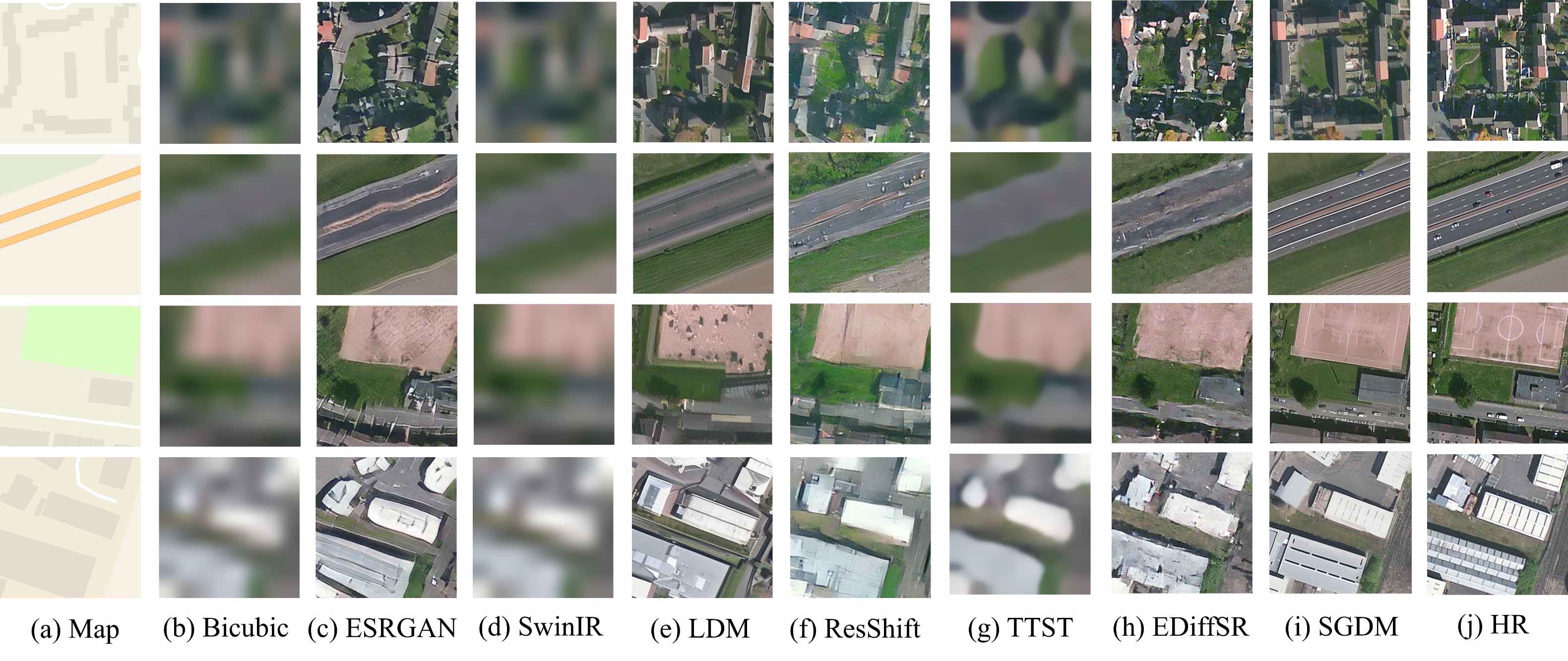}
    \caption{Visual comparison of 32$\times$ super-resolution on synthetic dataset. We select four representative scenes: residential areas, roads, squares, and factories.}
    \label{fig:sync_qualitative}
\end{figure*}

In Fig. \ref{fig:sync_qualitative}, we present a visual comparison of super-resolution results using different methods on the synthetic dataset. Due to the low spatial resolution of the LR input, a substantial amount of high-frequency information is lost and there exhibit significant semantic ambiguity. Consequently, traditional transformer-based regression models (SwinIR and TTST) can only produce overly smooth results, struggling to recover the lost high-frequency texture details. Among generative methods, ESRGAN and ResShift perform relatively better. ESRGAN can recover most of the high-frequency details, but suffer from serious distortion and deformation of the reconstructed ground objects. In contrast to ESRGAN, ResShift can reconstruct more regular shapes of objects, but it cannot recover as much high-frequency details as ESRGAN. LDM can generate richer and more realistic details compared to ESRGAN and ResShift. However, due to the lack of semantic guidance, positions and shapes of the generated objects exhibit some disarray. Thanks to the precise semantic guidance provided by vector maps and the powerful generative capabilities of Stable Diffusion, the proposed SGDM effectively addresses the problem of semantic ambiguity and the loss of high-frequency details, thereby achieving reconstruction results that balance realism and fidelity.

\begin{figure*}[htp]
    \centering
    \includegraphics[width=\textwidth]{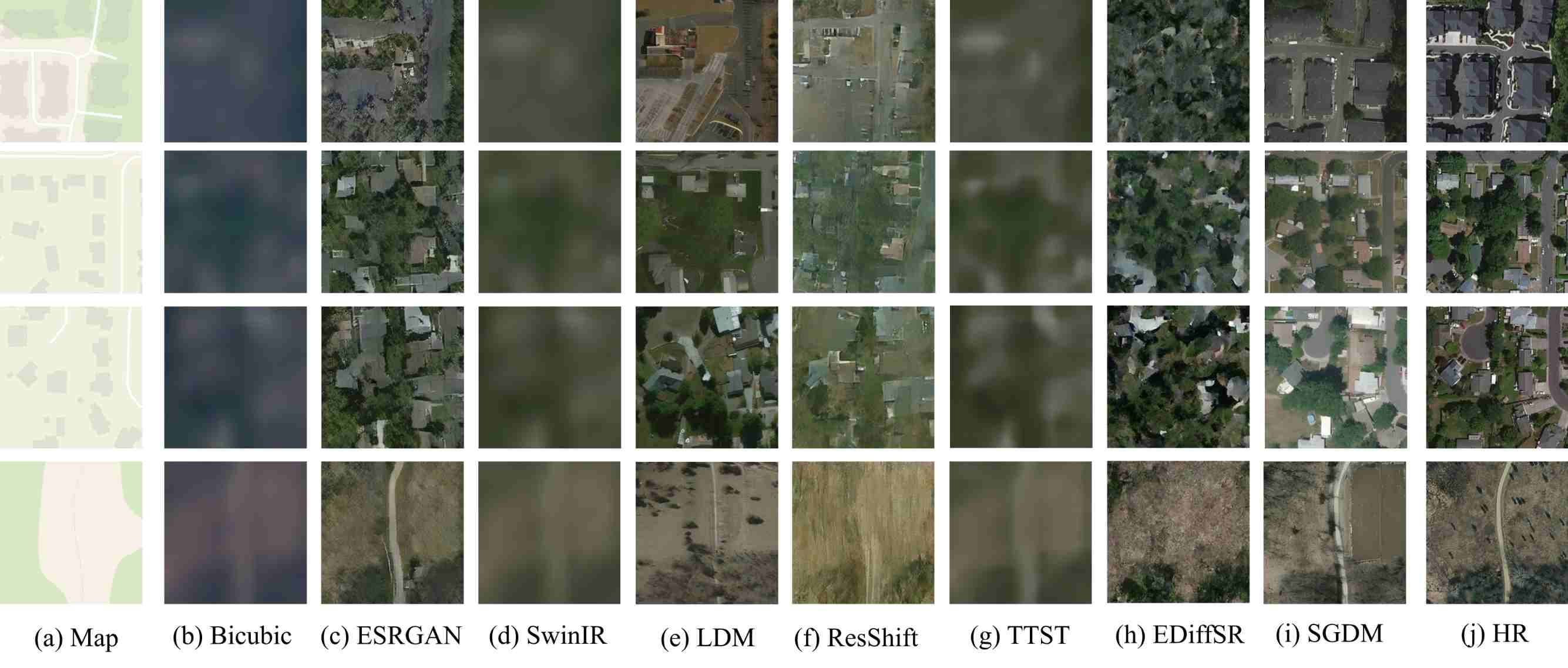}
    \caption{Visual comparison of 16$\times$ super-resolution on real-world dataset.}
    \label{fig:real_qualitative}
\end{figure*}

\subsection{Experiments on real-world data}\label{sec:Experiments on real-world data}
\subsubsection{Quantitative SR results}\label{sec:Quantitative results(real)}
Table \ref{tab:quantitative comparison} shows the evaluation results of different comparative methods on real-world data for 16$\times$ SR. Our proposed SGDM+ achieves the best performance on all no-reference metrics. However, it lags slightly behind ESRGAN in two full-reference perceptual metrics. This is because during testing, we do not provide any style guidance images and instead get completely random style vectors by sampling from the latent space of SFlow. Therefore, one-to-one evaluation metrics cannot fully demonstrate the advantages of SGDM+. However, the lower FID indicates that SGDM+ can better capture the distribution of HR images compared to other methods. This is because SGDM+ can independently model the style of HR through the SCM module, effectively overcoming the style misalignment between LR and HR in real-world scenarios.

\begin{figure*}[htp]
    \centering
    \includegraphics[width=\textwidth]{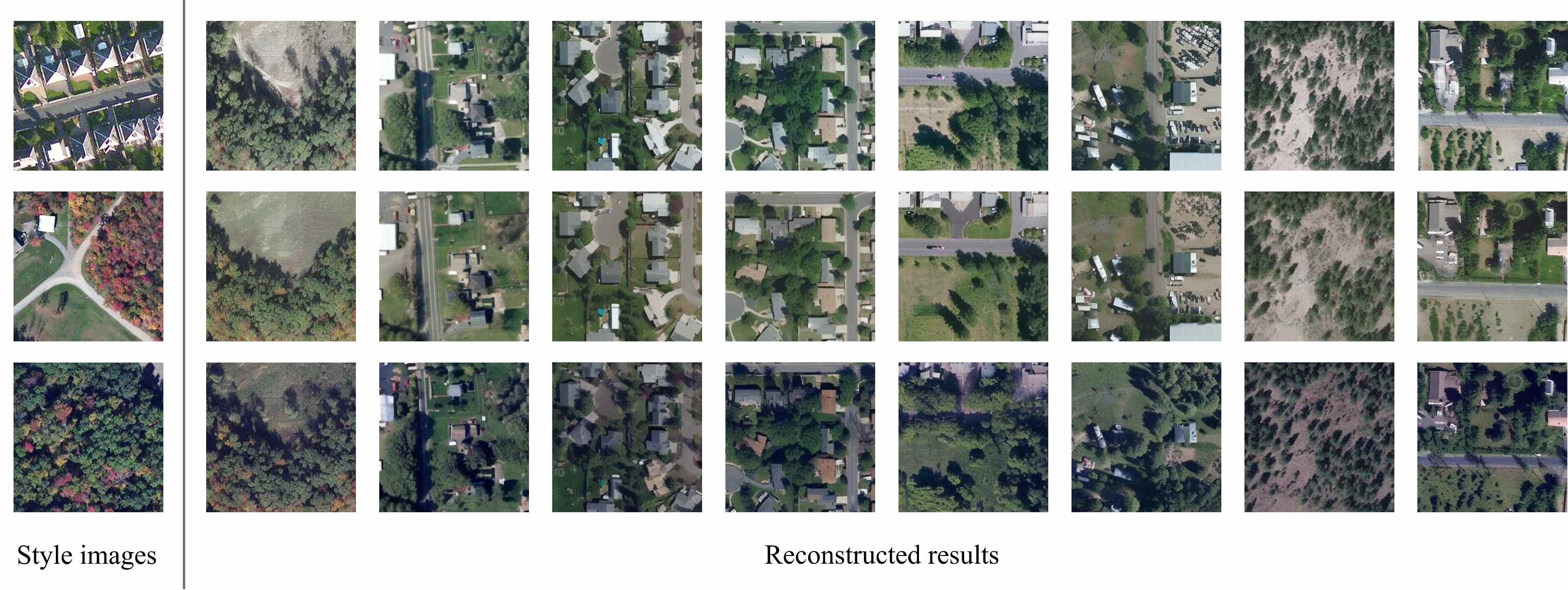}
    \caption{Visual comparison between reconstruction results given different style guidance images.}
    \label{fig:style_guidance}
\end{figure*}

\begin{figure*}[htp]
    \centering
    \includegraphics[width=\textwidth]{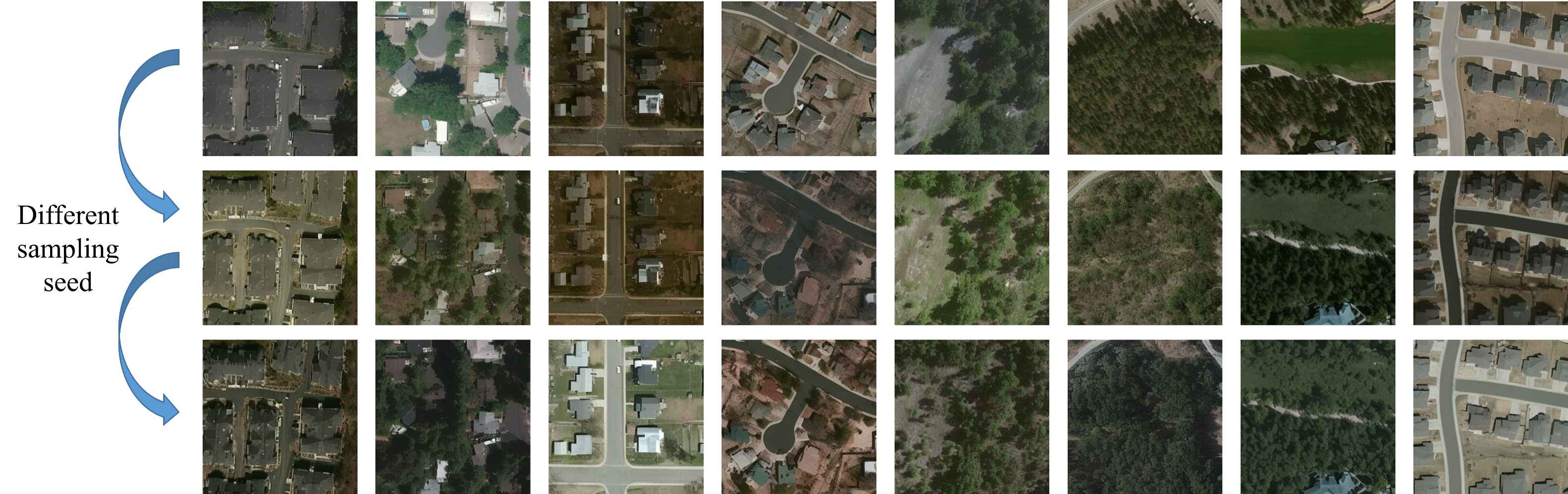}
    \caption{Visual comparison between reconstruction results with different sampling seeds for normalizing flow.}
    \label{fig:style_sample}
\end{figure*}

\subsubsection{Qualitative SR results}\label{sec:Qualitative results(real)}
Visual comparisons between different methods on real-world data are shown in Fig. \ref{fig:real_qualitative}. Due to the entanglement between content attributes and style attributes, the results provided by all comparative methods tend to have the average style of HR images, for example SR images tend to be darker. SGDM+ can perform style sampling based on the normalizing flow model, forming a more flexible solution space. Additionally, we find that due to the style differences between LR and HR in real-world datasets, some comparative methods even have difficulty converging during training (such as ResShift and EDiffSR), resulting in severely distorted SR results. ESRGAN, while capable of producing relatively accurate ground objects, falls short in generating rich textures. While LDM can generate results with higher realism, it often generates ground objects that do not match the ground truth due to the lack of accurate semantic guidance. SGDM+, on the other hand, obtains SR results that combine fidelity and realism under the guidance of vector maps and the support of pre-trained Stable Diffusion model.

\subsubsection{SR results of different style control strategies}\label{sec:Style control}
\textbf{Style guidance from example HR image.}
In many instances, addressing remote sensing image super-resolution often involves leveraging heterogeneous or historical remote sensing images as references. For example, \cite{dong2021rrsgan} proposed using HR remote sensing images captured by other satellite platforms as references to assist in the super-resolution process, thereby obtaining more realistic texture details. In contrast to their content-based reference methods, we advocate using HR remote sensing images as a style reference to overcome the style misalignment between LR and HR images. During inference, our framework can reuse the trained style encoder to extract the style vectors of the style guidance images, and then apply style injection on the LR feature maps through the AdaIN module. As shown in Fig. \ref{fig:style_guidance}, when providing different style guidance images, the model can perform instance-level style guidance (such as colors of buildings and roads, and the density of vegetation), thereby obtaining more controllable reconstruction results.

\textbf{Style sampled from fitted probabilistic distribution.}
In the most challenging scenario where no style guidance images are available, we can directly sample from the SFlow to acquire style vectors for super-resolution. It should be noted that due to the limited training data, the normalizing flow model can only approximate (rather than fully simulate) the style distribution of HR images. Therefore, compared to style guidance, style sampling will lower the quality and increase the uncertainty of the reconstruction results. As shown in Fig. \ref{fig:style_sample}, by changing the random seed for sampling from the style normalizing flow, we can obtain reconstruction results that are consistent in content but diverse in style. Due to the lack of appropriate style guidance, these results represent the lower bound of the performance of SGDM+ under extreme conditions (which is also the performance of SGDM+ recorded in Table \ref{tab:quantitative comparison}).

\begin{table*}
	\caption{Ablation study of our method (SGDM) on the synthetic split of CMSRD.}
	\label{tab:ablation_study}
	\centering
	\begin{tabular}{cccccc}
		\hline
         \makecell[c]{LR condition} &\makecell[c]{Vector map prior} &\makecell[c]{Stable Diffusion prior} &\makecell[c]{SPADE block} &LPIPS$\ \downarrow$ &FID$\ \downarrow$\cr
        \hline
        $\checkmark$ &$\checkmark$ &$\times$ &$\checkmark$ &0.8498 &277.2 \cr
        $\checkmark$ &$\times$ &$\checkmark$ &$\times$ &0.4652 &42.43 \cr
        $\checkmark$ &$\checkmark$ &$\checkmark$ &$\times$ &0.4478 &43.83 \cr
        $\checkmark$ &$\checkmark$ &$\checkmark$ &$\checkmark$ &\textbf{0.4344} &\textbf{36.63} \cr\hline
	\end{tabular}
\end{table*}

\begin{figure}[tp]
    \centering
    \includegraphics[width=\columnwidth]{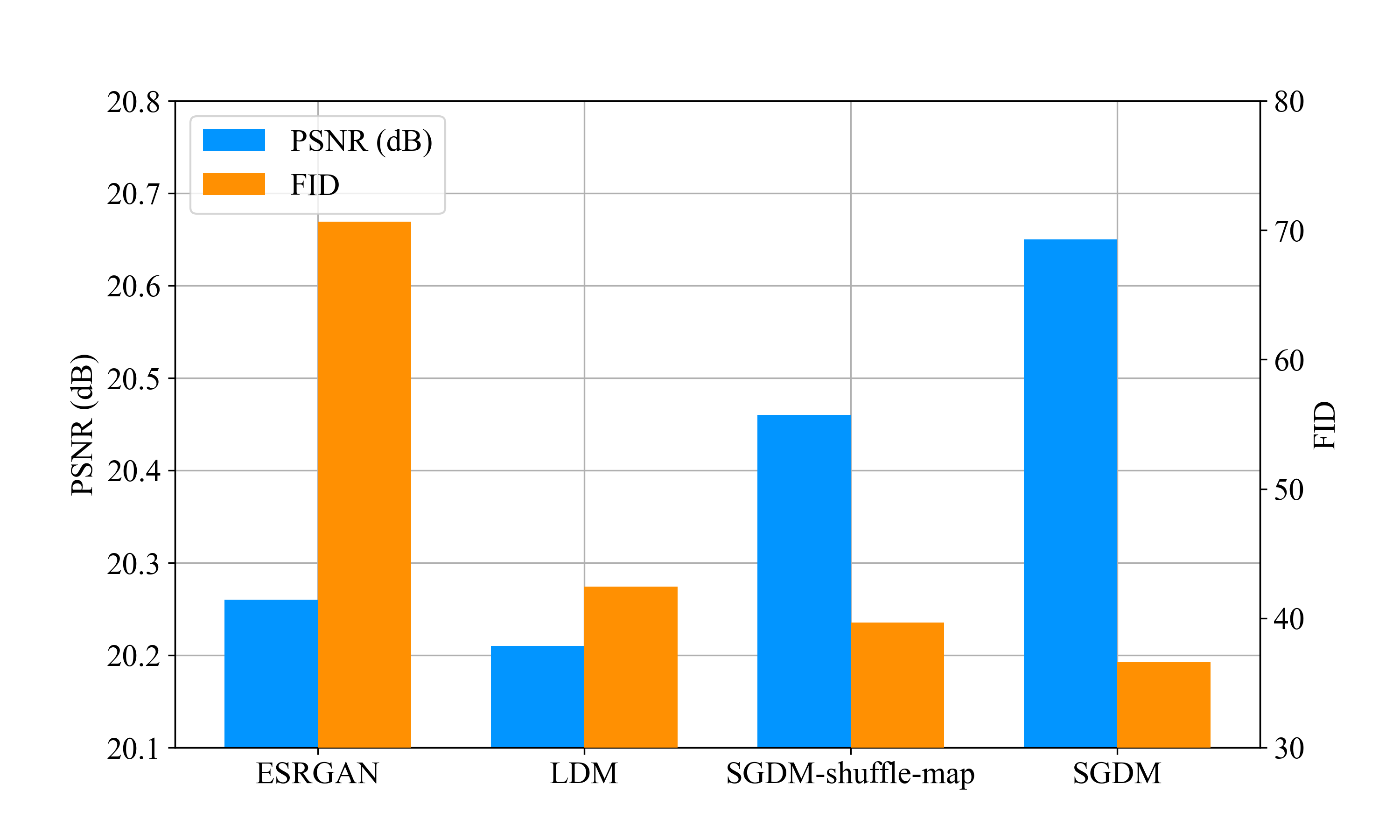}
    \caption{Performance of different methods on the synthetic dataset. `SGDM-shuffle-map' represents using vector maps completely unrelated to ground truth.}
    \label{fig:effectiveness-sg}
\end{figure}

\begin{figure}[tp]
    \centering
    \includegraphics[width=\columnwidth]{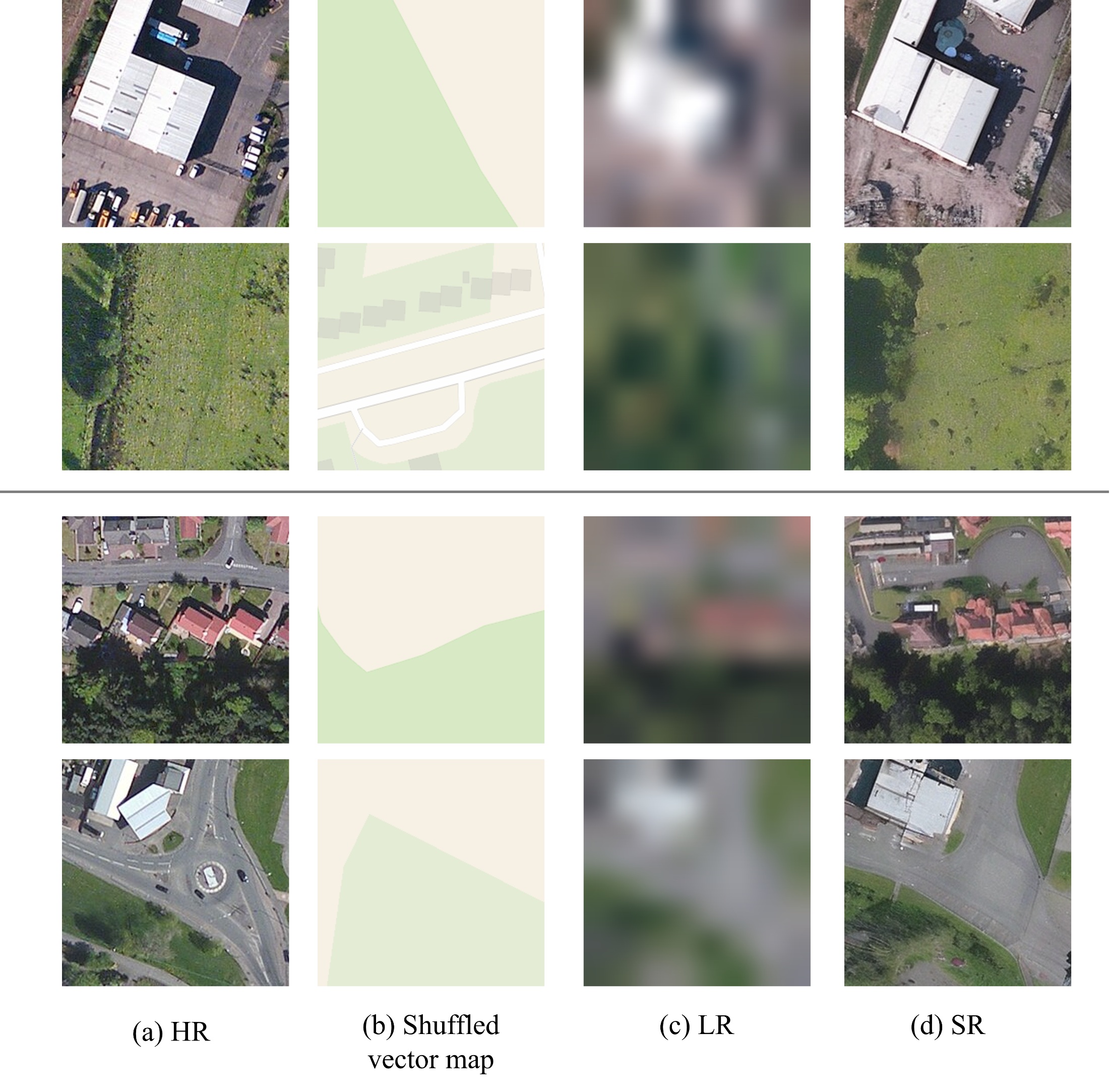}
    \caption{Visualization of the reconstruction results of SGDM given  vector map as semantic guidance. The reconstruction results of the first two rows are less affected by vector maps, while the last two rows are more significantly influenced.}
    \label{fig:sg-visual}
\end{figure}

\section{Ablation studies}\label{sec:Ablation studies}
\subsection{Effectiveness of diffusion prior}\label{sec:Effectiveness of diffusion prior}
The powerful generative capabilities of pre-trained Stable Diffusion can significantly enhance the visual quality of reconstruction outcomes. To assess its efficacy, we conduct an experiment where we employ our proposed content guided module (CGM) alone, bypassing the pre-trained Stable Diffusion as a prior. As demonstrated in the first row of Table \ref{tab:ablation_study}, the absence of the pre-trained Stable Diffusion prior leads to significantly lower perceptual quality in the reconstruction, due to the inherent limitations of traditional regression models in bridging the substantial spatial resolution disparity between LR and HR images.

\subsection{Effectiveness of semantic guidance}\label{sec:Effectiveness of semantic guidance}
For our proposed SGDM, vector maps serve as semantic guidance to enhance the fidelity of the reconstruction results. We evaluate the effectiveness of vector maps from two perspectives: whether vector maps are useful for improving the results, and whether the method we introduce vector maps is the most effective. For the former, we use only LR as the condition to guide the generation process of the diffusion model. For the latter, we eliminate all SPADE modules in the content guided module and simply merge the features of LR and the vector map by concatenation. The results of these experiments are presented in the second and third rows of Table \ref{tab:ablation_study}. These experiments demonstrate that incorporating vector maps through concatenation can also improve reconstruction fidelity. The SPADE module, as a spatially adaptive feature modulation technique, exploits the full potential of the semantic information from vector maps, thus yielding better results compared to simple concatenation.

Additionally, we also conduct some experiments to test the robustness of SGDM. By inputting vector maps that are entirely unrelated to the ground truth, we can assess the reliance of the model on vector maps. As depicted in Fig. \ref{fig:effectiveness-sg}, even with irrelevant vector maps, SGDM still generates relatively plausible reconstructions, outperforming LDM and ESRGAN. Conversely, when vector maps lack accurate semantic guidance, performance of the SGDM decreases, indicating the efficacy of vector map integration. From Fig. \ref{fig:sg-visual}, it can be observed that the reconstruction results of some remote sensing images, such as large scale buildings and cluttered texture areas, are not affected by the inaccurate semantic guidance from the vector maps. However, for dense urban scenes, the model heavily relies on semantic information provided by vector map, thereby leading to less satisfactory results.

\begin{figure}[tp]
    \centering
    \includegraphics[width=\columnwidth]{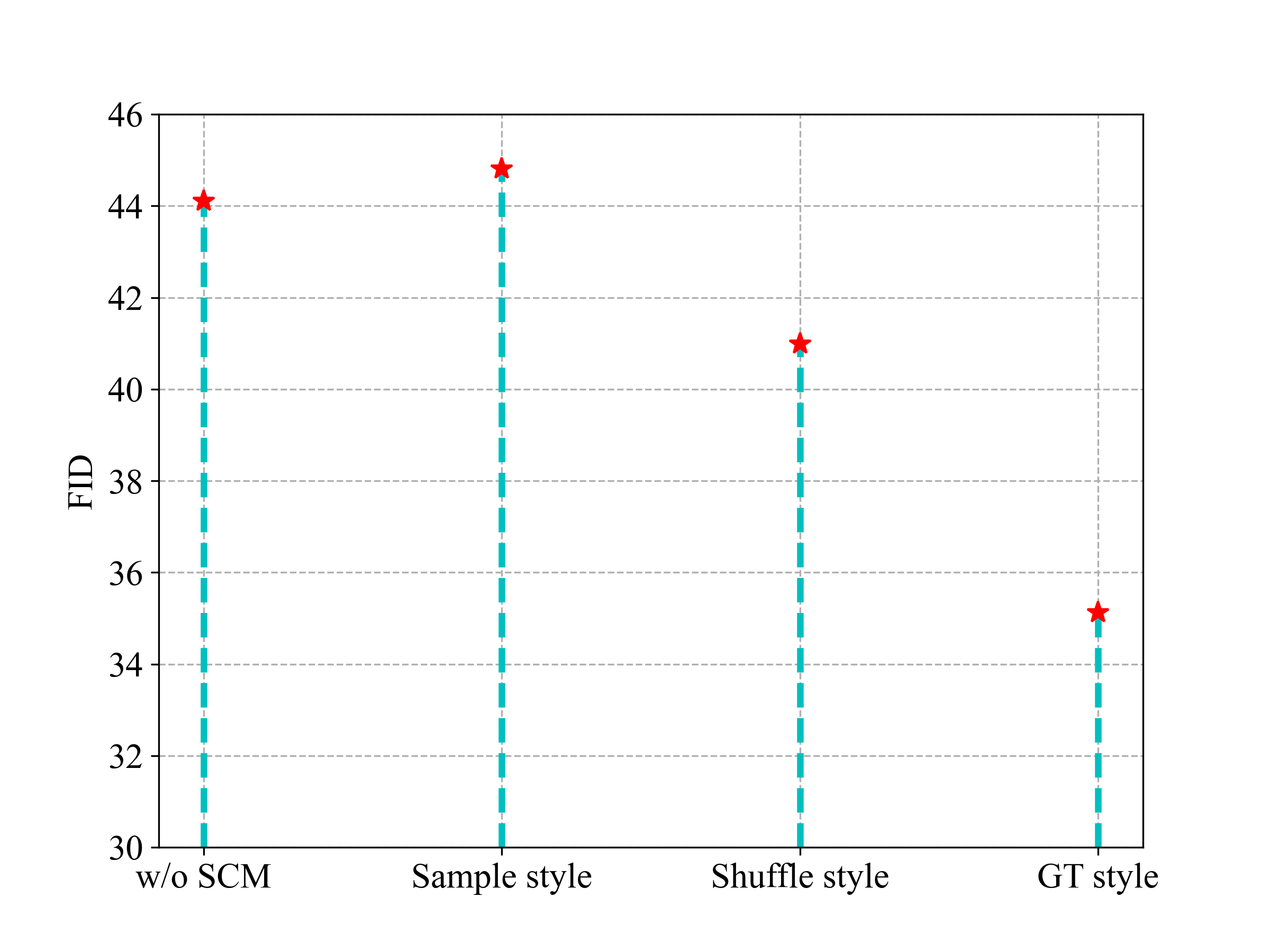}
    \caption{Performance of SGDM+ under different style guidance strategies. `Sample style' refers to style sampling through normalization flows. `Shuffle style' denotes style guidance using randomly chosen HR images. `GT style' indicates directly using the ground truth for style guidance.}
    \label{fig:effectiveness-scm}
\end{figure}

\begin{figure}[tp]
    \centering
    \includegraphics[width=\columnwidth]{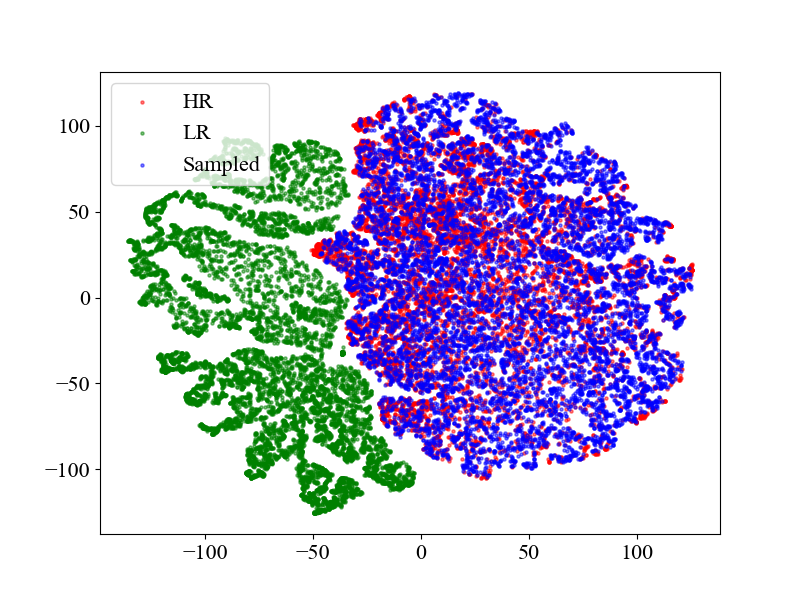}
    \caption{The t-SNE visualization of style vectors. It can be observed that there is a clear distribution difference between LR and HR, while our style normalization flow fits well into the style distribution of HR.}
    \label{fig:style_compare}
\end{figure}

\subsection{Effectiveness of style modelling}\label{sec:Effectiveness of style transfer}
To tackle the issue of style inconsistency between LR and HR images in real-world scenarios, we've developed a style correction module to enhance the capability of the model for style modelling. During testing, we can either employ the style encoder for style guidance or utilize normalization flow for random style sampling. Therefore, theoretically, completely random style sampling and using the ground truth for style guidance respectively represent the lower and upper bounds of the performance of SGDM+. Additionally, we can input randomly selected HR images for style guidance, which represents the average performance of SGDM+ under general conditions. We conduct these three inference methods separately and compare the results with the approach of not using the style correction module (\ie, SGDM). The obtained results are shown in Fig. \ref{fig:effectiveness-scm}. It can be observed that using the ground truth as style guidance and random style sampling achieve the best and worst reconstruction results. Notably, using randomly selected HR images for style guidance also produces favorable results, surpassing the model without the style correction module. These findings robustly attest to the efficacy and substantial practical applications of our proposed style correction module.

To visually inspect the distinct style distributions between LR and HR, we present a visualization of style vectors for 10,000 randomly chosen training pairs in Fig. \ref{fig:style_compare}. The comparison clearly illustrates a substantial difference in style distribution between HR and LR (denoted as `HR style space' and `LR style space' in Fig. \ref{fig:motivation}). Our proposed style normalization flow effectively aligns with the HR image style distribution. During testing, by directly sampling from the normalization flow, we can provide useful style priors to the model, thereby enhancing the reconstruction fidelity.

\begin{figure*}[tp]
    \centering
    \includegraphics[width=\textwidth]{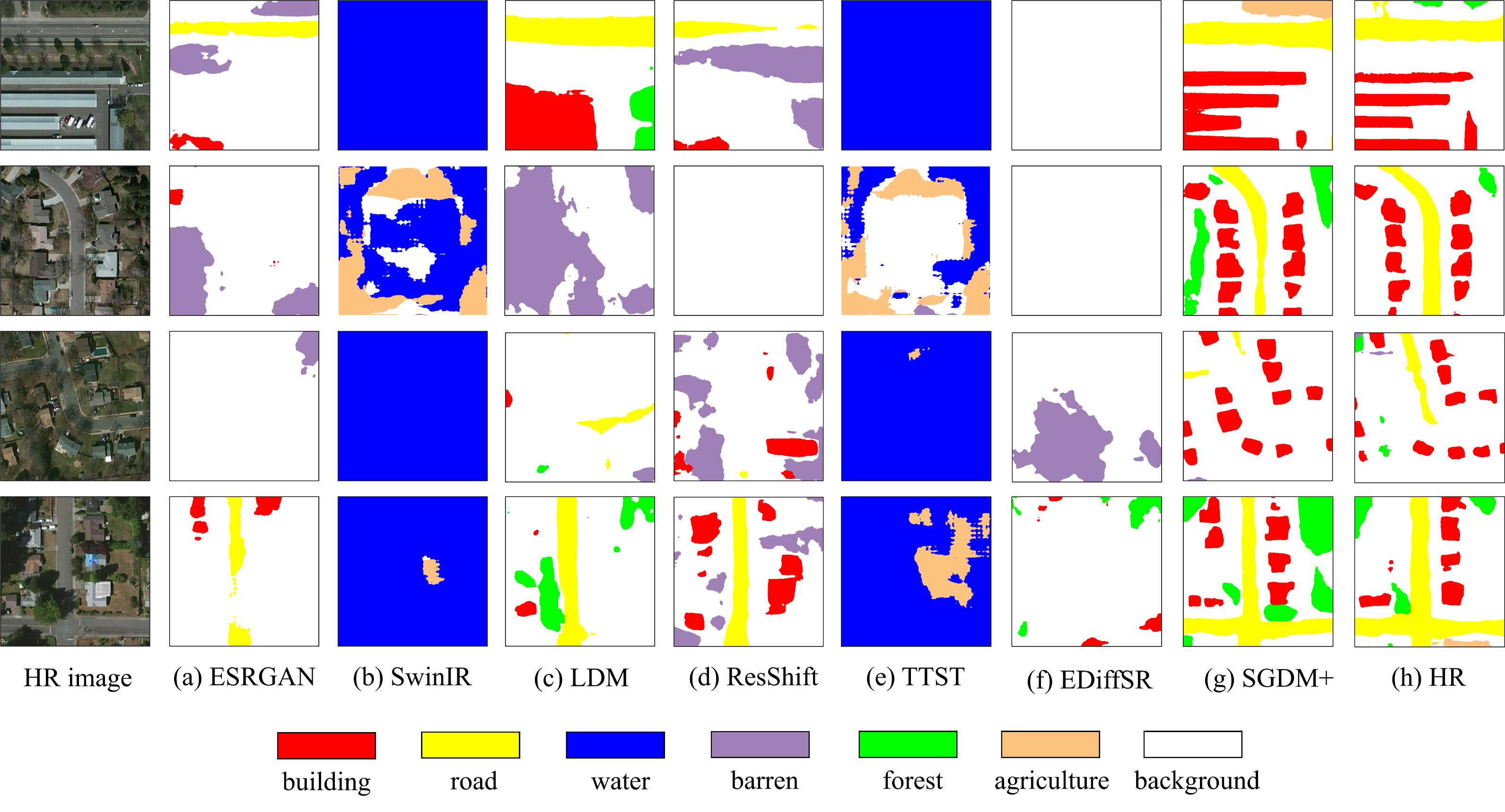}
    \caption{Visual comparison of semantic segmentation results from different super-resolution methods.}
    \label{fig:task_segment}
\end{figure*}

\begin{figure}[tp]
    \centering
    \includegraphics[width=\columnwidth]{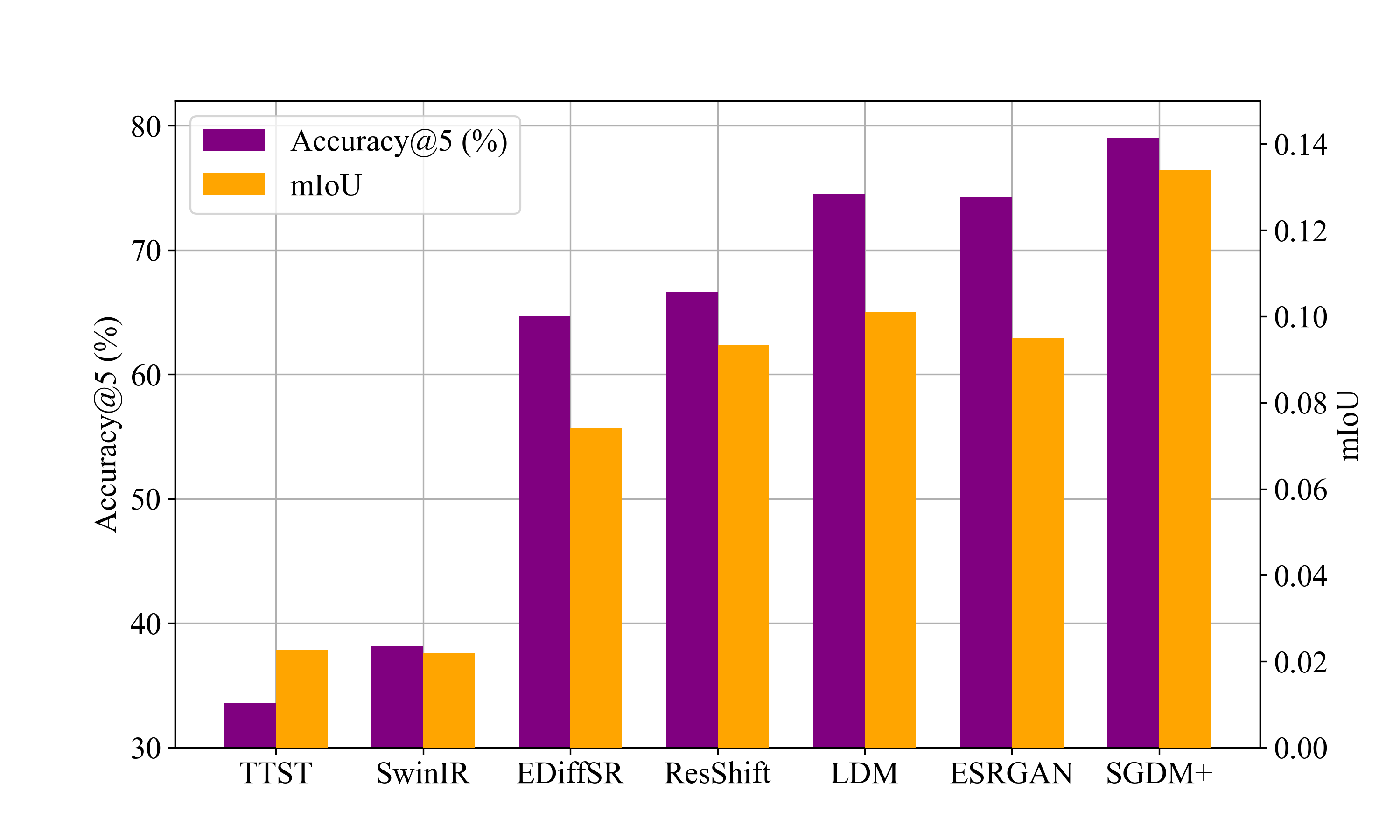}
    \caption{Quantitative comparison of scene recognition and semantic segmentation results among different super-resolution methods.}
    \label{fig:task_metric}
\end{figure}

\subsection{Evaluate usability of SR images on downstream vision tasks}\label{sec:downstream_tasks}
The evaluation of natural image quality primarily relies on quantitative metrics and visual appeal. However, the value of remote sensing images lies in their performance on relevant analysis tasks. To assess the reconstruction quality of various SR methods comprehensively, we choose two representative downstream tasks (scene recognition and semantic segmentation) to showcase the superiority of our proposed method in perceptual quality. Due to the absence of the ground truth, we leverage a pre-trained remote sensing foundation model (RVSA \citep{wang2022advancing}) with strong generalization capabilities to analyze the SR results. Our focus is on the relative performance of SR results compared to HR images in the context of these tasks, not the absolute accuracy of SR results. Thus all metrics, such as classification accuracy and mean intersection over union (mIoU), are computed using HR analysis results as a reference.

\textbf{Scene recognition.}
The scene recognition task aims to classify input remote sensing images, enabling a comprehensive assessment of global semantic consistency between SR results and HR. We employ the RVSA model fine-tuned on the UC Merced Land-Use dataset \citep{yang2010bag} to classify the test set images in the real-world CMSRD dataset. As shown in Fig. \ref{fig:task_metric}, our proposed SGDM+ achieves the highest classification accuracy (\ie, the most similar classification results compared to HR), with a top-5 accuracy close to 80\%. This indicates that the SR results of SGDM+ have the highest fidelity for machine perception. Notably, all generative models, including SGDM+, outperform regression models like TTST and SwinIR, highlighting the limitations of pixel-level loss for machine vision tasks.

\textbf{Semantic segmentation.}
The semantic segmentation task aims to classify each pixel of a given remote sensing image. We utilize this task to evaluate the consistency between SR results and HR in terms of local semantics. We use the RVSA model fine-tuned on the LoveDA dataset \citep{wang2021loveda} to segment the test images, and the mean intersection over union (mIoU) of segmentation results is displayed in Fig. \ref{fig:task_metric}. Consistent with scene recognition, SGDM+ delivers the best segmentation outcomes, significantly outperforming other methods. In Fig. \ref{fig:task_segment}, visual comparisons reveal that most comparative methods struggle with precise ground object reconstruction, while SGDM+ provides the closest results to HR, particularly for buildings and roads, underscoring the promising prospects of our proposed method.

\section{Conclusion}\label{sec:Conclusion}
In this paper, we propose a novel framework (SGDM) for large scale factor remote sensing image super-resolution. To address challenges such as semantic inaccuracies and texture blurriness commonly encountered in this task, we utilize vector maps to provide semantic guidance and exploit the powerful generative capabilities of the pre-trained Stable Diffusion model to generate more realistic SR images. Additionally, considering the style differences between LR and HR caused by the sensor-specific imaging characteristics in real-world scenarios, we develop a method for extracting style information and modeling its probability distribution. During testing, diverse outputs can be obtained through style guidance or style sampling.

Extensive experiments are conducted on our proposed CMSRD dataset. Both qualitative and quantitative experiments demonstrate the superiority of our proposed framework. Comprehensive ablation experiments validate the necessity of key components in SGDM, including vector maps, generative priors, and style correction module. Finally, we further demonstrate the application prospects of our proposed framework through two downstream tasks: scene recognition and semantic segmentation. In future work, we will further explore the feasibility of large scale factor remote sensing image super-resolution by introducing richer explicit priors (such as digital elevation maps, historical remote sensing images, etc.) and higher-quality generative priors (such as pre-trained models for remote sensing images).

{
    \small
    \bibliographystyle{ieeenat_fullname}
    \bibliography{main}

\begin{thebibliography}{69}
\providecommand{\natexlab}[1]{#1}
\providecommand{\url}[1]{\texttt{#1}}
\expandafter\ifx\csname urlstyle\endcsname\relax
  \providecommand{\doi}[1]{doi: #1}\else
  \providecommand{\doi}{doi: \begingroup \urlstyle{rm}\Url}\fi

\bibitem[Bamford et~al.(2020)Bamford, Kelly, Dalla~Rosa, Cade, Fretwell,
  Trathan, Cubaynes, Mesquita, Gerrish, Friedlaender,
  et~al.]{bamford2020comparison}
CCG Bamford, N Kelly, L Dalla~Rosa, DE Cade, PT Fretwell, PN Trathan, HC
  Cubaynes, AFC Mesquita, L Gerrish, AS Friedlaender, et~al.
\newblock A comparison of baleen whale density estimates derived from
  overlapping satellite imagery and a shipborne survey.
\newblock \emph{Scientific reports}, 10\penalty0 (1):\penalty0 12985, 2020.

\bibitem[Buhler et~al.(2020)Buhler, Romero, and Timofte]{buhler2020deepsee}
Marcel~C Buhler, Andr{\'e}s Romero, and Radu Timofte.
\newblock Deepsee: Deep disentangled semantic explorative extreme
  super-resolution.
\newblock In \emph{Proceedings of the Asian Conference on Computer Vision},
  2020.

\bibitem[Bulat and Tzimiropoulos(2018)]{bulat2018super}
Adrian Bulat and Georgios Tzimiropoulos.
\newblock Super-fan: Integrated facial landmark localization and
  super-resolution of real-world low resolution faces in arbitrary poses with
  gans.
\newblock In \emph{Proceedings of the IEEE conference on computer vision and
  pattern recognition}, pages 109--117, 2018.

\bibitem[Chan et~al.(2021)Chan, Wang, Xu, Gu, and Loy]{chan2021glean}
Kelvin~CK Chan, Xintao Wang, Xiangyu Xu, Jinwei Gu, and Chen~Change Loy.
\newblock Glean: Generative latent bank for large-factor image
  super-resolution.
\newblock In \emph{Proceedings of the IEEE/CVF conference on computer vision
  and pattern recognition}, pages 14245--14254, 2021.

\bibitem[Ding et~al.(2020)Ding, Ma, Wang, and Simoncelli]{ding2020image}
Keyan Ding, Kede Ma, Shiqi Wang, and Eero~P Simoncelli.
\newblock Image quality assessment: Unifying structure and texture similarity.
\newblock \emph{IEEE transactions on pattern analysis and machine
  intelligence}, 44\penalty0 (5):\penalty0 2567--2581, 2020.

\bibitem[Dinh et~al.(2016)Dinh, Sohl-Dickstein, and Bengio]{dinh2016density}
Laurent Dinh, Jascha Sohl-Dickstein, and Samy Bengio.
\newblock Density estimation using real nvp.
\newblock \emph{arXiv preprint arXiv:1605.08803}, 2016.

\bibitem[Dong et~al.(2015)Dong, Loy, He, and Tang]{dong2015image}
Chao Dong, Chen~Change Loy, Kaiming He, and Xiaoou Tang.
\newblock Image super-resolution using deep convolutional networks.
\newblock \emph{IEEE transactions on pattern analysis and machine
  intelligence}, 38\penalty0 (2):\penalty0 295--307, 2015.

\bibitem[Dong et~al.(2021)Dong, Zhang, and Fu]{dong2021rrsgan}
Runmin Dong, Lixian Zhang, and Haohuan Fu.
\newblock Rrsgan: Reference-based super-resolution for remote sensing image.
\newblock \emph{IEEE Transactions on Geoscience and Remote Sensing},
  60:\penalty0 1--17, 2021.

\bibitem[Dong et~al.(2024)Dong, Yuan, Luo, Chen, Zhang, Zhang, Li, Zheng, and
  Fu]{dong2024building}
Runmin Dong, Shuai Yuan, Bin Luo, Mengxuan Chen, Jinxiao Zhang, Lixian Zhang,
  Weijia Li, Juepeng Zheng, and Haohuan Fu.
\newblock Building bridges across spatial and temporal resolutions:
  Reference-based super-resolution via change priors and conditional diffusion
  model.
\newblock \emph{arXiv preprint arXiv:2403.17460}, 2024.

\bibitem[Freeman et~al.(2002)Freeman, Jones, and Pasztor]{freeman2002example}
William~T Freeman, Thouis~R Jones, and Egon~C Pasztor.
\newblock Example-based super-resolution.
\newblock \emph{IEEE Computer graphics and Applications}, 22\penalty0
  (2):\penalty0 56--65, 2002.

\bibitem[Gu et~al.(2020)Gu, Shen, and Zhou]{gu2020image}
Jinjin Gu, Yujun Shen, and Bolei Zhou.
\newblock Image processing using multi-code gan prior.
\newblock In \emph{Proceedings of the IEEE/CVF conference on computer vision
  and pattern recognition}, pages 3012--3021, 2020.

\bibitem[Heusel et~al.(2017)Heusel, Ramsauer, Unterthiner, Nessler, and
  Hochreiter]{heusel2017gans}
Martin Heusel, Hubert Ramsauer, Thomas Unterthiner, Bernhard Nessler, and Sepp
  Hochreiter.
\newblock Gans trained by a two time-scale update rule converge to a local nash
  equilibrium.
\newblock \emph{Advances in neural information processing systems}, 30, 2017.

\bibitem[Ho et~al.(2020)Ho, Jain, and Abbeel]{ho2020denoising}
Jonathan Ho, Ajay Jain, and Pieter Abbeel.
\newblock Denoising diffusion probabilistic models.
\newblock \emph{Advances in neural information processing systems},
  33:\penalty0 6840--6851, 2020.

\bibitem[Huang and Belongie(2017)]{huang2017arbitrary}
Xun Huang and Serge Belongie.
\newblock Arbitrary style transfer in real-time with adaptive instance
  normalization.
\newblock In \emph{Proceedings of the IEEE international conference on computer
  vision}, pages 1501--1510, 2017.

\bibitem[Johnson et~al.(2016)Johnson, Alahi, and
  Fei-Fei]{johnson2016perceptual}
Justin Johnson, Alexandre Alahi, and Li Fei-Fei.
\newblock Perceptual losses for real-time style transfer and super-resolution.
\newblock In \emph{Computer Vision--ECCV 2016: 14th European Conference,
  Amsterdam, The Netherlands, October 11-14, 2016, Proceedings, Part II 14},
  pages 694--711. Springer, 2016.

\bibitem[Karras et~al.(2019)Karras, Laine, and Aila]{karras2019style}
Tero Karras, Samuli Laine, and Timo Aila.
\newblock A style-based generator architecture for generative adversarial
  networks.
\newblock In \emph{Proceedings of the IEEE/CVF conference on computer vision
  and pattern recognition}, pages 4401--4410, 2019.

\bibitem[Karras et~al.(2020)Karras, Laine, Aittala, Hellsten, Lehtinen, and
  Aila]{karras2020analyzing}
Tero Karras, Samuli Laine, Miika Aittala, Janne Hellsten, Jaakko Lehtinen, and
  Timo Aila.
\newblock Analyzing and improving the image quality of stylegan.
\newblock In \emph{Proceedings of the IEEE/CVF conference on computer vision
  and pattern recognition}, pages 8110--8119, 2020.

\bibitem[Ke et~al.(2021)Ke, Wang, Wang, Milanfar, and Yang]{ke2021musiq}
Junjie Ke, Qifei Wang, Yilin Wang, Peyman Milanfar, and Feng Yang.
\newblock Musiq: Multi-scale image quality transformer.
\newblock In \emph{Proceedings of the IEEE/CVF international conference on
  computer vision}, pages 5148--5157, 2021.

\bibitem[Lai et~al.(2017)Lai, Huang, Ahuja, and Yang]{lai2017deep}
Wei-Sheng Lai, Jia-Bin Huang, Narendra Ahuja, and Ming-Hsuan Yang.
\newblock Deep laplacian pyramid networks for fast and accurate
  super-resolution.
\newblock In \emph{Proceedings of the IEEE conference on computer vision and
  pattern recognition}, pages 624--632, 2017.

\bibitem[Ledig et~al.(2017)Ledig, Theis, Husz{\'a}r, Caballero, Cunningham,
  Acosta, Aitken, Tejani, Totz, Wang, et~al.]{ledig2017photo}
Christian Ledig, Lucas Theis, Ferenc Husz{\'a}r, Jose Caballero, Andrew
  Cunningham, Alejandro Acosta, Andrew Aitken, Alykhan Tejani, Johannes Totz,
  Zehan Wang, et~al.
\newblock Photo-realistic single image super-resolution using a generative
  adversarial network.
\newblock In \emph{Proceedings of the IEEE conference on computer vision and
  pattern recognition}, pages 4681--4690, 2017.

\bibitem[Li et~al.(2020)Li, Wan, Cheng, Meng, and Han]{li2020object}
Ke Li, Gang Wan, Gong Cheng, Liqiu Meng, and Junwei Han.
\newblock Object detection in optical remote sensing images: A survey and a new
  benchmark.
\newblock \emph{ISPRS journal of photogrammetry and remote sensing},
  159:\penalty0 296--307, 2020.

\bibitem[Li et~al.(2023)Li, Ren, Jin, Lan, Wang, Zeng, Wang, and
  Chen]{li2023diffusion}
Xin Li, Yulin Ren, Xin Jin, Cuiling Lan, Xingrui Wang, Wenjun Zeng, Xinchao
  Wang, and Zhibo Chen.
\newblock Diffusion models for image restoration and enhancement--a
  comprehensive survey.
\newblock \emph{arXiv preprint arXiv:2308.09388}, 2023.

\bibitem[Li et~al.(2017)Li, Fang, Yang, Wang, Lu, and Yang]{li2017universal}
Yijun Li, Chen Fang, Jimei Yang, Zhaowen Wang, Xin Lu, and Ming-Hsuan Yang.
\newblock Universal style transfer via feature transforms.
\newblock \emph{Advances in neural information processing systems}, 30, 2017.

\bibitem[Liang et~al.(2021{\natexlab{a}})Liang, Cao, Sun, Zhang, Van~Gool, and
  Timofte]{liang2021swinir}
Jingyun Liang, Jiezhang Cao, Guolei Sun, Kai Zhang, Luc Van~Gool, and Radu
  Timofte.
\newblock Swinir: Image restoration using swin transformer.
\newblock In \emph{Proceedings of the IEEE/CVF international conference on
  computer vision}, pages 1833--1844, 2021{\natexlab{a}}.

\bibitem[Liang et~al.(2021{\natexlab{b}})Liang, Lugmayr, Zhang, Danelljan,
  Van~Gool, and Timofte]{liang2021hierarchical}
Jingyun Liang, Andreas Lugmayr, Kai Zhang, Martin Danelljan, Luc Van~Gool, and
  Radu Timofte.
\newblock Hierarchical conditional flow: A unified framework for image
  super-resolution and image rescaling.
\newblock In \emph{Proceedings of the IEEE/CVF International Conference on
  Computer Vision}, pages 4076--4085, 2021{\natexlab{b}}.

\bibitem[Liao et~al.(2023)Liao, Xiao, Yang, Ma, Wang, and Satoh]{liao2023high}
Liang Liao, Jing Xiao, Yan Yang, Xujie Ma, Zheng Wang, and Shin’ichi Satoh.
\newblock High temporal frequency vehicle counting from low-resolution
  satellite images.
\newblock \emph{ISPRS Journal of Photogrammetry and Remote Sensing},
  198:\penalty0 45--59, 2023.

\bibitem[Liebel and K{\"o}rner(2016)]{liebel2016single}
Lukas Liebel and Marco K{\"o}rner.
\newblock Single-image super resolution for multispectral remote sensing data
  using convolutional neural networks.
\newblock \emph{The International Archives of the Photogrammetry, Remote
  Sensing and Spatial Information Sciences}, 41:\penalty0 883--890, 2016.

\bibitem[Lim et~al.(2017)Lim, Son, Kim, Nah, and Mu~Lee]{lim2017enhanced}
Bee Lim, Sanghyun Son, Heewon Kim, Seungjun Nah, and Kyoung Mu~Lee.
\newblock Enhanced deep residual networks for single image super-resolution.
\newblock In \emph{Proceedings of the IEEE conference on computer vision and
  pattern recognition workshops}, pages 136--144, 2017.

\bibitem[Lin et~al.(2023)Lin, He, Chen, Lyu, Fei, Dai, Ouyang, Qiao, and
  Dong]{lin2023diffbir}
Xinqi Lin, Jingwen He, Ziyan Chen, Zhaoyang Lyu, Ben Fei, Bo Dai, Wanli Ouyang,
  Yu Qiao, and Chao Dong.
\newblock Diffbir: Towards blind image restoration with generative diffusion
  prior.
\newblock \emph{arXiv preprint arXiv:2308.15070}, 2023.

\bibitem[Liu et~al.(2021)Liu, Siu, and Chan]{liu:2021photo}
Zhi-Song Liu, Wan-Chi Siu, and Yui-Lam Chan.
\newblock Photo-realistic image super-resolution via variational autoencoders.
\newblock \emph{IEEE Transactions on Circuits and Systems for Video
  Technology}, 31\penalty0 (4):\penalty0 1351--1365, 2021.

\bibitem[Lugmayr et~al.(2020)Lugmayr, Danelljan, Van~Gool, and
  Timofte]{lugmayr2020srflow}
Andreas Lugmayr, Martin Danelljan, Luc Van~Gool, and Radu Timofte.
\newblock Srflow: Learning the super-resolution space with normalizing flow.
\newblock In \emph{Computer Vision--ECCV 2020: 16th European Conference,
  Glasgow, UK, August 23--28, 2020, Proceedings, Part V 16}, pages 715--732.
  Springer, 2020.

\bibitem[Meng et~al.(2022)Meng, Li, Lei, Zou, and Shi]{meng2022large}
Yapeng Meng, Wenyuan Li, Sen Lei, Zhengxia Zou, and Zhenwei Shi.
\newblock Large-factor super-resolution of remote sensing images with
  spectra-guided generative adversarial networks.
\newblock \emph{IEEE Transactions on Geoscience and Remote Sensing},
  60:\penalty0 1--11, 2022.

\bibitem[Menon et~al.(2020)Menon, Damian, Hu, Ravi, and Rudin]{menon2020pulse}
Sachit Menon, Alexandru Damian, Shijia Hu, Nikhil Ravi, and Cynthia Rudin.
\newblock Pulse: Self-supervised photo upsampling via latent space exploration
  of generative models.
\newblock In \emph{Proceedings of the ieee/cvf conference on computer vision
  and pattern recognition}, pages 2437--2445, 2020.

\bibitem[Park et~al.(2019)Park, Liu, Wang, and Zhu]{park2019semantic}
Taesung Park, Ming-Yu Liu, Ting-Chun Wang, and Jun-Yan Zhu.
\newblock Semantic image synthesis with spatially-adaptive normalization.
\newblock In \emph{Proceedings of the IEEE/CVF conference on computer vision
  and pattern recognition}, pages 2337--2346, 2019.

\bibitem[Pineda et~al.(2020)Pineda, Ayma, and Beltran]{pineda2020generative}
F Pineda, V Ayma, and C Beltran.
\newblock A generative adversarial network approach for super-resolution of
  sentinel-2 satellite images.
\newblock \emph{The International Archives of the Photogrammetry, Remote
  Sensing and Spatial Information Sciences}, 43:\penalty0 9--14, 2020.

\bibitem[Rombach et~al.(2022)Rombach, Blattmann, Lorenz, Esser, and
  Ommer]{rombach2022high}
Robin Rombach, Andreas Blattmann, Dominik Lorenz, Patrick Esser, and Bj{\"o}rn
  Ommer.
\newblock High-resolution image synthesis with latent diffusion models.
\newblock In \emph{Proceedings of the IEEE/CVF conference on computer vision
  and pattern recognition}, pages 10684--10695, 2022.

\bibitem[Saharia et~al.(2022)Saharia, Ho, Chan, Salimans, Fleet, and
  Norouzi]{saharia2022image}
Chitwan Saharia, Jonathan Ho, William Chan, Tim Salimans, David~J Fleet, and
  Mohammad Norouzi.
\newblock Image super-resolution via iterative refinement.
\newblock \emph{IEEE Transactions on Pattern Analysis and Machine
  Intelligence}, 45\penalty0 (4):\penalty0 4713--4726, 2022.

\bibitem[Shaham et~al.(2019)Shaham, Dekel, and Michaeli]{shaham2019singan}
Tamar~Rott Shaham, Tali Dekel, and Tomer Michaeli.
\newblock Singan: Learning a generative model from a single natural image.
\newblock In \emph{Proceedings of the IEEE/CVF international conference on
  computer vision}, pages 4570--4580, 2019.

\bibitem[Sun et~al.(2019)Sun, Bian, Zhou, and Pan]{sun2019using}
Chuanliang Sun, Yan Bian, Tao Zhou, and Jianjun Pan.
\newblock Using of multi-source and multi-temporal remote sensing data improves
  crop-type mapping in the subtropical agriculture region.
\newblock \emph{Sensors}, 19\penalty0 (10):\penalty0 2401, 2019.

\bibitem[Sun et~al.(2008)Sun, Xu, and Shum]{sun2008image}
Jian Sun, Zongben Xu, and Heung-Yeung Shum.
\newblock Image super-resolution using gradient profile prior.
\newblock In \emph{2008 IEEE conference on computer vision and pattern
  recognition}, pages 1--8. IEEE, 2008.

\bibitem[Sun and Chen(2022)]{sun2022learning}
Wanjie Sun and Zhenzhong Chen.
\newblock Learning discrete representations from reference images for large
  scale factor image super-resolution.
\newblock \emph{IEEE Transactions on Image Processing}, 31:\penalty0
  1490--1503, 2022.

\bibitem[Timofte et~al.(2013)Timofte, De~Smet, and
  Van~Gool]{timofte2013anchored}
Radu Timofte, Vincent De~Smet, and Luc Van~Gool.
\newblock Anchored neighborhood regression for fast example-based
  super-resolution.
\newblock In \emph{Proceedings of the IEEE international conference on computer
  vision}, pages 1920--1927, 2013.

\bibitem[Tong et~al.(2020)Tong, Xia, Lu, Shen, Li, You, and
  Zhang]{tong2020land}
Xin-Yi Tong, Gui-Song Xia, Qikai Lu, Huanfeng Shen, Shengyang Li, Shucheng You,
  and Liangpei Zhang.
\newblock Land-cover classification with high-resolution remote sensing images
  using transferable deep models.
\newblock \emph{Remote Sensing of Environment}, 237:\penalty0 111322, 2020.

\bibitem[Ulyanov et~al.(2016)Ulyanov, Vedaldi, and
  Lempitsky]{ulyanov2016instance}
Dmitry Ulyanov, Andrea Vedaldi, and Victor Lempitsky.
\newblock Instance normalization: The missing ingredient for fast stylization.
\newblock \emph{arXiv preprint arXiv:1607.08022}, 2016.

\bibitem[Venter et~al.(2020)Venter, Brousse, Esau, and
  Meier]{venter2020hyperlocal}
Zander~S Venter, Oscar Brousse, Igor Esau, and Fred Meier.
\newblock Hyperlocal mapping of urban air temperature using remote sensing and
  crowdsourced weather data.
\newblock \emph{Remote Sensing of Environment}, 242:\penalty0 111791, 2020.

\bibitem[Wang et~al.(2022)Wang, Zhang, Xu, Zhang, Du, Tao, and
  Zhang]{wang2022advancing}
Di Wang, Qiming Zhang, Yufei Xu, Jing Zhang, Bo Du, Dacheng Tao, and Liangpei
  Zhang.
\newblock Advancing plain vision transformer toward remote sensing foundation
  model.
\newblock \emph{IEEE Transactions on Geoscience and Remote Sensing},
  61:\penalty0 1--15, 2022.

\bibitem[Wang et~al.(2021)Wang, Zheng, Ma, Lu, and Zhong]{wang2021loveda}
Junjue Wang, Zhuo Zheng, Ailong Ma, Xiaoyan Lu, and Yanfei Zhong.
\newblock Loveda: A remote sensing land-cover dataset for domain adaptive
  semantic segmentation.
\newblock \emph{arXiv preprint arXiv:2110.08733}, 2021.

\bibitem[Wang et~al.(2023{\natexlab{a}})Wang, Chan, and Loy]{wang2023exploring}
Jianyi Wang, Kelvin~CK Chan, and Chen~Change Loy.
\newblock Exploring clip for assessing the look and feel of images.
\newblock In \emph{Proceedings of the AAAI Conference on Artificial
  Intelligence}, pages 2555--2563, 2023{\natexlab{a}}.

\bibitem[Wang et~al.(2023{\natexlab{b}})Wang, Yue, Zhou, Chan, and
  Loy]{wang2023exploiting}
Jianyi Wang, Zongsheng Yue, Shangchen Zhou, Kelvin~CK Chan, and Chen~Change
  Loy.
\newblock Exploiting diffusion prior for real-world image super-resolution.
\newblock \emph{arXiv preprint arXiv:2305.07015}, 2023{\natexlab{b}}.

\bibitem[Wang et~al.(2018{\natexlab{a}})Wang, Yu, Wu, Gu, Liu, Dong, Qiao, and
  Change~Loy]{wang2018esrgan}
Xintao Wang, Ke Yu, Shixiang Wu, Jinjin Gu, Yihao Liu, Chao Dong, Yu Qiao, and
  Chen Change~Loy.
\newblock Esrgan: Enhanced super-resolution generative adversarial networks.
\newblock In \emph{Proceedings of the European conference on computer vision
  (ECCV) workshops}, pages 0--0, 2018{\natexlab{a}}.

\bibitem[Wang et~al.(2018{\natexlab{b}})Wang, Perazzi, McWilliams,
  Sorkine-Hornung, Sorkine-Hornung, and Schroers]{wang2018fully}
Yifan Wang, Federico Perazzi, Brian McWilliams, Alexander Sorkine-Hornung, Olga
  Sorkine-Hornung, and Christopher Schroers.
\newblock A fully progressive approach to single-image super-resolution.
\newblock In \emph{Proceedings of the IEEE conference on computer vision and
  pattern recognition workshops}, pages 864--873, 2018{\natexlab{b}}.

\bibitem[Wang et~al.(2004)Wang, Bovik, Sheikh, and Simoncelli]{wang2004image}
Zhou Wang, Alan~C Bovik, Hamid~R Sheikh, and Eero~P Simoncelli.
\newblock Image quality assessment: from error visibility to structural
  similarity.
\newblock \emph{IEEE transactions on image processing}, 13\penalty0
  (4):\penalty0 600--612, 2004.

\bibitem[Xiao et~al.(2023)Xiao, Yuan, Jiang, He, Jin, and
  Zhang]{xiao2023ediffsr}
Yi Xiao, Qiangqiang Yuan, Kui Jiang, Jiang He, Xianyu Jin, and Liangpei Zhang.
\newblock Ediffsr: An efficient diffusion probabilistic model for remote
  sensing image super-resolution.
\newblock \emph{IEEE Transactions on Geoscience and Remote Sensing}, 2023.

\bibitem[Xiao et~al.(2024)Xiao, Yuan, Jiang, He, Lin, and Zhang]{xiao2024ttst}
Yi Xiao, Qiangqiang Yuan, Kui Jiang, Jiang He, Chia-Wen Lin, and Liangpei
  Zhang.
\newblock Ttst: A top-k token selective transformer for remote sensing image
  super-resolution.
\newblock \emph{IEEE Transactions on Image Processing}, 2024.

\bibitem[Yang et~al.(2010)Yang, Wright, Huang, and Ma]{yang2010image}
Jianchao Yang, John Wright, Thomas~S Huang, and Yi Ma.
\newblock Image super-resolution via sparse representation.
\newblock \emph{IEEE transactions on image processing}, 19\penalty0
  (11):\penalty0 2861--2873, 2010.

\bibitem[Yang et~al.(2023{\natexlab{a}})Yang, Zhang, Song, Hong, Xu, Zhao,
  Zhang, Cui, and Yang]{yang2023diffusion}
Ling Yang, Zhilong Zhang, Yang Song, Shenda Hong, Runsheng Xu, Yue Zhao, Wentao
  Zhang, Bin Cui, and Ming-Hsuan Yang.
\newblock Diffusion models: A comprehensive survey of methods and applications.
\newblock \emph{ACM Computing Surveys}, 56\penalty0 (4):\penalty0 1--39,
  2023{\natexlab{a}}.

\bibitem[Yang et~al.(2023{\natexlab{b}})Yang, Ren, Xie, and
  Zhang]{yang2023pixel}
Tao Yang, Peiran Ren, Xuansong Xie, and Lei Zhang.
\newblock Pixel-aware stable diffusion for realistic image super-resolution and
  personalized stylization.
\newblock \emph{arXiv preprint arXiv:2308.14469}, 2023{\natexlab{b}}.

\bibitem[Yang et~al.(2017)Yang, Feng, Yang, Zhao, Liu, Guo, and
  Yan]{yang2017deep}
Wenhan Yang, Jiashi Feng, Jianchao Yang, Fang Zhao, Jiaying Liu, Zongming Guo,
  and Shuicheng Yan.
\newblock Deep edge guided recurrent residual learning for image
  super-resolution.
\newblock \emph{IEEE Transactions on Image Processing}, 26\penalty0
  (12):\penalty0 5895--5907, 2017.

\bibitem[Yang and Newsam(2010)]{yang2010bag}
Yi Yang and Shawn Newsam.
\newblock Bag-of-visual-words and spatial extensions for land-use
  classification.
\newblock In \emph{Proceedings of the 18th SIGSPATIAL international conference
  on advances in geographic information systems}, pages 270--279, 2010.

\bibitem[Ye et~al.(2023)Ye, Zhang, Liu, Han, and Yang]{ye2023ip}
Hu Ye, Jun Zhang, Sibo Liu, Xiao Han, and Wei Yang.
\newblock Ip-adapter: Text compatible image prompt adapter for text-to-image
  diffusion models.
\newblock \emph{arXiv preprint arXiv:2308.06721}, 2023.

\bibitem[Yin et~al.(2021)Yin, Dong, Hamm, Li, Wang, Xing, and
  Fu]{yin2021integrating}
Jiadi Yin, Jinwei Dong, Nicholas~AS Hamm, Zhichao Li, Jianghao Wang, Hanfa
  Xing, and Ping Fu.
\newblock Integrating remote sensing and geospatial big data for urban land use
  mapping: A review.
\newblock \emph{International Journal of Applied Earth Observation and
  Geoinformation}, 103:\penalty0 102514, 2021.

\bibitem[Yue et~al.(2024)Yue, Wang, and Loy]{yue2024resshift}
Zongsheng Yue, Jianyi Wang, and Chen~Change Loy.
\newblock Resshift: Efficient diffusion model for image super-resolution by
  residual shifting.
\newblock \emph{Advances in Neural Information Processing Systems}, 36, 2024.

\bibitem[Zhang et~al.(2021)Zhang, Liang, Van~Gool, and
  Timofte]{zhang2021designing}
Kai Zhang, Jingyun Liang, Luc Van~Gool, and Radu Timofte.
\newblock Designing a practical degradation model for deep blind image
  super-resolution.
\newblock In \emph{Proceedings of the IEEE/CVF International Conference on
  Computer Vision}, pages 4791--4800, 2021.

\bibitem[Zhang et~al.(2023)Zhang, Rao, and Agrawala]{zhang2023adding}
Lvmin Zhang, Anyi Rao, and Maneesh Agrawala.
\newblock Adding conditional control to text-to-image diffusion models.
\newblock In \emph{Proceedings of the IEEE/CVF International Conference on
  Computer Vision}, pages 3836--3847, 2023.

\bibitem[Zhang et~al.(2018{\natexlab{a}})Zhang, Isola, Efros, Shechtman, and
  Wang]{zhang2018unreasonable}
Richard Zhang, Phillip Isola, Alexei~A Efros, Eli Shechtman, and Oliver Wang.
\newblock The unreasonable effectiveness of deep features as a perceptual
  metric.
\newblock In \emph{Proceedings of the IEEE conference on computer vision and
  pattern recognition}, pages 586--595, 2018{\natexlab{a}}.

\bibitem[Zhang et~al.(2020{\natexlab{a}})Zhang, Yuan, Li, Sun, and
  Zhang]{zhang2020scene}
Shu Zhang, Qiangqiang Yuan, Jie Li, Jing Sun, and Xuguo Zhang.
\newblock Scene-adaptive remote sensing image super-resolution using a
  multiscale attention network.
\newblock \emph{IEEE Transactions on Geoscience and Remote Sensing},
  58\penalty0 (7):\penalty0 4764--4779, 2020{\natexlab{a}}.

\bibitem[Zhang et~al.(2018{\natexlab{b}})Zhang, Li, Li, Wang, Zhong, and
  Fu]{zhang2018image}
Yulun Zhang, Kunpeng Li, Kai Li, Lichen Wang, Bineng Zhong, and Yun Fu.
\newblock Image super-resolution using very deep residual channel attention
  networks.
\newblock In \emph{Proceedings of the European conference on computer vision
  (ECCV)}, pages 286--301, 2018{\natexlab{b}}.

\bibitem[Zhang et~al.(2020{\natexlab{b}})Zhang, Zhang, DiVerdi, Wang,
  Echevarria, and Fu]{zhang2020texture}
Yulun Zhang, Zhifei Zhang, Stephen DiVerdi, Zhaowen Wang, Jose Echevarria, and
  Yun Fu.
\newblock Texture hallucination for large-factor painting super-resolution.
\newblock In \emph{European Conference on Computer Vision}, pages 209--225.
  Springer, 2020{\natexlab{b}}.

\bibitem[Zhu et~al.(2020)Zhu, Talebi, Shi, Yang, and Milanfar]{zhu2020super}
Xiang Zhu, Hossein Talebi, Xinwei Shi, Feng Yang, and Peyman Milanfar.
\newblock Super-resolving commercial satellite imagery using realistic training
  data.
\newblock In \emph{2020 IEEE International Conference on Image Processing
  (ICIP)}, pages 498--502. IEEE, 2020.

\end{thebibliography}
}


\end{document}